\documentclass{article}

\usepackage[final,main]{neurips_2026}

\usepackage[utf8]{inputenc}
\usepackage[T1]{fontenc}
\usepackage{float}
\usepackage{wrapfig}
\usepackage{graphicx}
\usepackage{hyperref}
\usepackage{url}
\usepackage{booktabs}
\usepackage{multirow}
\usepackage{amsmath}
\usepackage{amsfonts}
\usepackage{nicefrac}
\usepackage{microtype}
\usepackage{xcolor}
\usepackage{enumitem}

\newcommand{\ci}[3]{#1{\scriptsize\,[#2,#3]}}

\newif\ifshowcodexnotes
\showcodexnotesfalse

\newcommand{\codex}[1]{\ifshowcodexnotes\textcolor{blue}{#1}\fi}

\title{Emergent Steering Beyond Endpoint Alignment in Chemical Reaction Models}

\author{%
  Yili Shen \quad Kehan Guo \quad Haomin Zhuang \quad Xiangliang Zhang \\
  University of Notre Dame \\
  \texttt{\{yshen7, kguo2, hzhuang2, xzhang33\}@nd.edu} \\
}

\begin{document}

\maketitle

\begin{abstract}
Machine learning models for chemical reaction prediction are usually evaluated by endpoint accuracy, but endpoint correctness alone does not reveal how a model represents chemical transformation or why it fails. We introduce \textbf{UniRxnAxis}, a unified forward--retrosynthesis framework that models both tasks as direction-conditioned endpoint redistribution in a shared latent space. By aligning source states, transformed states, and encoded target endpoints, UniRxnAxis enables directional auditing of learned latent updates. This shared latent geometry allows 
the learned displacement (from the encoder source state \(S_0\) to the decoder-facing transformed state \(S_1\))  to be decomposed into an endpoint-parallel component and an endpoint-orthogonal residual, which we call \emph{emergent steering}. Our evaluation shows that emergent steering is functionally important: in retrosynthesis, it recovers most of the model's utility, remains effective under random, shuffled, and prototype controls, and directly affects which bond edits are promoted or suppressed. The residual-over-parallel ordering persists across alignment strengths, after decoder-interface calibration, against angle-matched random axes, and in a second graph-edit architecture, although its magnitude depends on the reference axis and optimization. The decomposition also supports a proxy taxonomy of model-side failures, showing how reaction models can be audited beyond endpoint accuracy.

\end{abstract}

\section{Introduction}

Machine learning models have been widely applied to chemical reaction modeling by translating between molecular representations, such as SMILES strings or molecular graphs \citep{coley2019graph,wan2022retroformer,shi2020g2gs,zhong2023graph2edits,chen2023g2retro}. In this paradigm, models map reactants to products for forward reaction prediction, or products to reactants for retrosynthesis. In both cases, the central goal is to \textbf{model the underlying chemical transformation between the input and output molecules}, including which bonds are broken, which bonds are formed, and how atoms are rearranged during the reaction.
On standard benchmarks, forward reaction prediction has achieved high  accuracy (measured by comparing model outputs with the recorded ground truth) \citep{bi2021non,joung2025electron}. In contrast, retrosynthesis remains inherently ambiguous \citep{fortunato2020data,tu2022permutation,bradshaw2025challenging}: multiple valid reaction routes can lead to the same product, yet datasets typically record only a single target, making exact-match evaluation underrepresent true performance.

Does this apparent progress 
mean that machine learning models are ready for routine use by chemists? Not quite. Despite strong benchmark results, these models are still largely perceived as black boxes. Chemists are often reluctant to rely on predictions that lack transparency or cannot be interrogated in a scientifically meaningful way.
Importantly, this limitation goes beyond standard notions of \emph{interpretability}, which often rely on post hoc analyses such as attention weights, gradient-based attribution, or input perturbations 
\citep{kovacs2021quantitative,hastedt2024investigating}. While useful, these methods do not fully capture the underlying transformation process.

For practical adoption in chemistry,  models must   provide more than endpoint correctness. 
Concretely, they should  expose audit handles that allow chemists to inspect promoted or suppressed bond edits, assess whether computation concentrates near plausible reaction centers, and diagnose where transformations fail. In this paper, we develop models that support such auditing, enabling chemists to interrogate and validate predicted transformations rather than passively accept final endpoints.

Since many models use an encoder--decoder framework, we study chemical transformations in latent space. Given an encoded molecule $S_0$, corresponding to either reactants or a product, the model maps it to a transformed representation $S_1$,   which is then decoded into the output molecule (product in forward prediction or the reactants in retrosynthesis). Let 
$\Delta z=S_1-S_0$. 
We can ask whether $\Delta z$ can be decomposed into 
 chemically meaningful components whose roles can be tested through decoder behavior, bond-edit decisions, and reaction-center localization.

We propose to decompose $\Delta z$ into two components: a target-aligned component that captures the intended chemical transformation, and a complementary orthogonal component that captures \emph{emergent steering}:
\begin{equation}
\Delta z = \Delta z_{\parallel} + \Delta z_{\perp},
\end{equation}
where $\Delta z_{\parallel}$ explains the  coarse commitment motion, and $\Delta z_{\perp}$ represents  additional latent-space shifts that may steer the model toward chemically plausible  sample-specific outcomes, conceptually shown in Fig.~\ref{fig:directional_decomposition_intro}.
To define the decomposition, we use the encoded ground-truth endpoint as a reference.
For a source state \(S_0\), let \(g = z_{\mathrm{target}} - S_0\) be the supervised endpoint direction.
We define \(\Delta z_{\parallel}\) as the projection of \(\Delta z\) onto \(g\), and \(\Delta z_{\perp}\) as the remaining component orthogonal to \(g\).

To make these latent quantities comparable and the decomposition measurable, we introduce UniRxnAxis: Unified Reaction Modeling with Directional Latent Axes, which is a joint forward--retrosynthesis framework with bidirectional endpoint alignment. UniRxnAxis uses a shared encoder to place reactant and product endpoints in a common latent coordinate system, a direction-conditioned transition module with task adapters to distinguish forward prediction from retrosynthesis, and a source-conditioned residual decoder to reconstruct the target endpoint from the source state and learned latent update. This design serves not only endpoint prediction, but also auditability: it makes the source state \(S_0\), transformed state \(S_1\), and encoded target endpoint \(z_{\mathrm{target}}\) directly comparable. As a result, \(g=z_{\mathrm{target}}-S_0\) defines a target-reference axis for analyzing the realized update \(\Delta z=S_1-S_0\).


\IfFileExists{figures/directional_decomposition_functional_asymmetry.pdf}{%
\begin{figure}[t]
\centering
\includegraphics[width=0.8\textwidth]{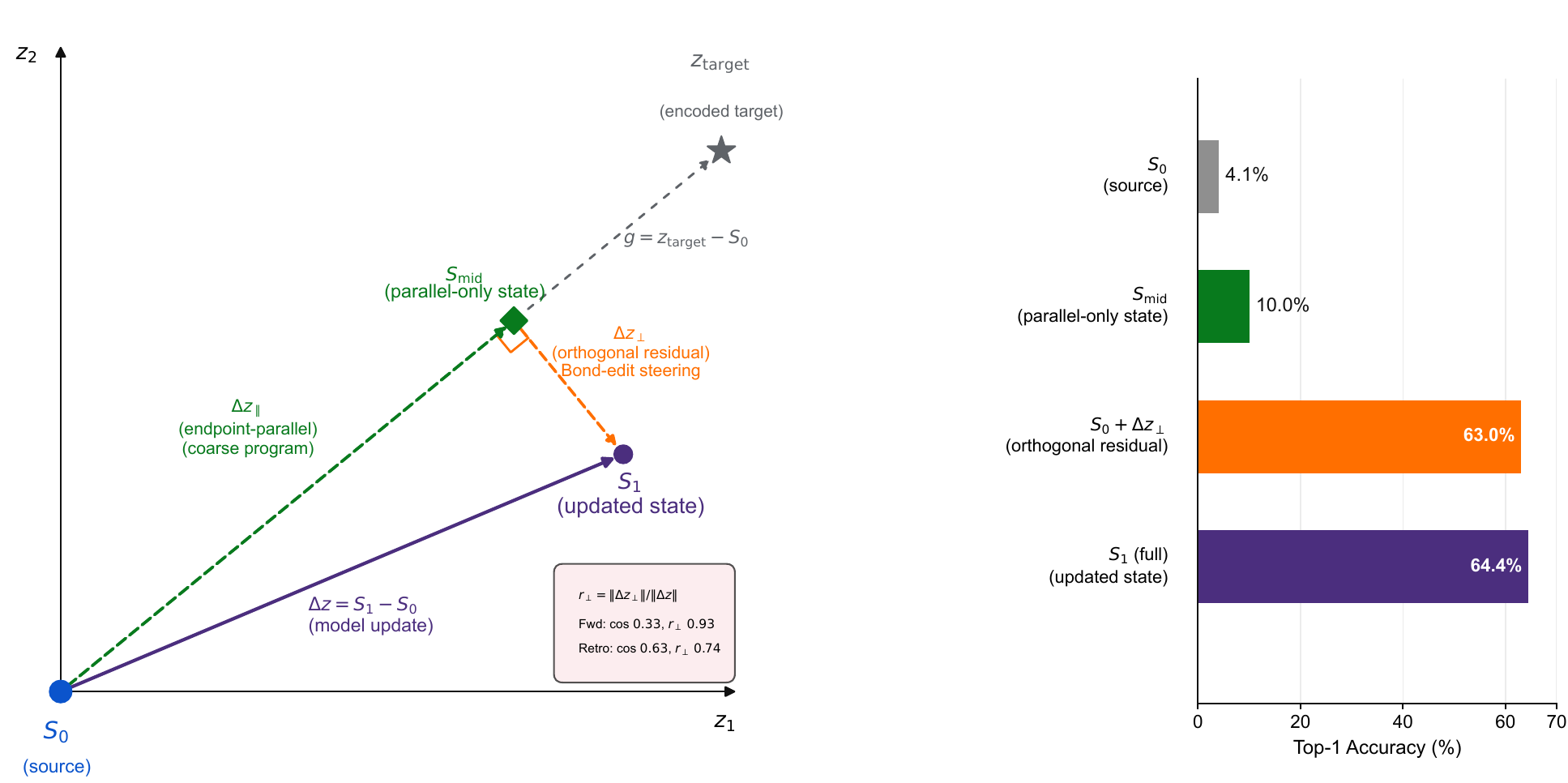}
\caption{Directional decomposition and functional asymmetry. (Left) The model
update is decomposed into endpoint-parallel content $\Delta z_{\parallel}$ and an
orthogonal residual $\Delta z_{\perp}$ relative to
$g=z_{\mathrm{target}}-S_0$. (Right) In retrosynthesis, 
endpoint-parallel decoding collapses ($S_{mid}$) to 10\% top-1 accuracy, while the orthogonal residual ($S_0 + \Delta z_{\perp}$) preserves nearly full utility (63.0\% vs. 64.4\%).} \vspace{-0.1in}
\label{fig:directional_decomposition_intro}
\end{figure}
}{}

Our evaluation shows that \(\Delta z_{\perp}\) recovers most retrosynthesis utility, survives control substitutions, and influences which bond edits are promoted or suppressed.
Because the encoded target is itself weakly decoder-executable, part of this asymmetry reflects the choice of reference axis; the gap shrinks but does not close when the axis is made more executable (Sec.~\ref{sec:axis_dependence}). As \(\Delta z_{\perp}\) uses the ground-truth endpoint, emergent steering is an audit-time diagnostic, not an inference-time control.

To summarize, our contributions are:
\begin{itemize}[leftmargin=*] \vspace{-0.1in}
\item We introduce UniRxnAxis, a unified forward--retrosynthesis framework that formulates both tasks as direction-conditioned endpoint redistribution in a shared latent coordinate system.
\item We propose a directional decomposition of the learned latent update into an endpoint-parallel component and an endpoint-orthogonal residual (\emph{emergent steering}), providing an audit handle for model-side reaction steering.
\item Extensive evaluation shows that emergent steering is functionally important: \(\Delta z_{\perp}\)
preserves most retrosynthesis performance, remains robust to control substitutions, and modulates bond-edit decisions. We characterize how the effect depends on the reference axis, the alignment weight, and the decoder interface, and reproduce its qualitative ordering in a second architecture (Graph2Edits). The same decomposition also enables a proxy taxonomy of model-side failures. 
\end{itemize}

\vspace{-0.2in}
\section{Related Work} \vspace{-0.1in}

\textbf{Reaction Prediction and Retrosynthesis.}
Reaction prediction and retrosynthesis are commonly formulated as encoder--decoder problems. Sequence-to-sequence~\citep{schwaller2018found,schwaller2019molecular,tetko2020state,irwin2022chemformer,pesciullesi2020transfer,lu2022unified,wang2020heck,zhang2021data,liu2017retrosynthetic,karpov2019transformer,wan2022retroformer}, graph-to-graph~\citep{jin2017predicting,coley2019graph,do2019graph,sacha2021molecule,bi2021non,meng2023doubly,shi2020g2gs,somnath2021graphretro,zhong2023graph2edits,chen2023g2retro,zhao2025retromtgr,yao2024node}, template-based~\citep{segler2017neuralsym,coley2017retrosim,dai2019gln,chen2021localretro,xie2023retrosynthesis}, template-free~\citep{liu2017retrosynthetic,karpov2019transformer,tetko2020state,irwin2022chemformer,tu2022permutation,wan2022retroformer}, and hybrid systems~\citep{tu2022permutation} have all made this framing effective: given one side of a reaction, the model is trained to recover the recorded other side.  
Prior methods have sought to improve latent alignment between input and output molecules, primarily as a means to enhance endpoint prediction accuracy ~\citep{bradshaw2019model,kim2021valid,li2025retromoe,shen2026driving}.  
In contrast, our work treats latent alignment as a tool for analyzing the transformation itself. 
Prior bidirectional models use forward and retrosynthetic signals mainly to improve reactant--product mapping at the molecular level~\cite{lu2022unified,xu2025unified,lin2025synbridge}. In contrast, we use bidirectional modeling not only for prediction, but also to characterize latent steering. 

\textbf{Interpretability of reaction models.}  
Typical post hoc probes of trained reaction-translation models have been used to explain  the predictions.  For example,  Integrated Gradients is employed to attribute the predictions of a GCN-based retrosynthesis ~\citep{ishida2019prediction}. 
Similarly, T5-Chem~\citep{lu2022unified}  uses SHAP~\citep{lundberg2017shap} values to explain the importance of functional groups in reaction tasks. Also, attribution analysis is employed in interpreting the trained transformer's dependency on input tokens of a reaction. ~\citep{kovacs2021quantitative}
These methods are useful for explaining which input features or molecular regions influence a prediction, answering which input features contributed to this output. However, our goal is different: UniRxnAxis audits the model's internal transformation process 
and identify which part of that move is decoder-executable.
Moreover,  our work is distinct from standard mechanistic interpretability. In chemistry, ``mechanism'' usually refers to arrow-pushing steps, reactive structures, or isolable intermediates, and we do not claim to recover these directly. Instead, we study the model mechanism: a functionally consequential organization of latent computation that is aligned with reaction-relevant structure.  


\vspace{-0.05in}
\section{Methodology} 
\label{sec:method}
\vspace{-0.1in}

\textbf{Problem formulation: reaction as endpoint redistribution.}
We formulate reaction modeling as role-conditioned endpoint redistribution. Molecular reactions transform one molecular graph into another through bond formation, bond breaking, and atom rearrangement. Thus, forward prediction maps the reactant endpoint to the product endpoint, while retrosynthesis maps the product endpoint to the reactant endpoint.
Let an endpoint molecular graph be $G=(\mathcal V,\mathcal E,Q,A)$, where
$\mathcal V$ denotes atom slots, $\mathcal E$ bond orders, $Q$ formal charges,
and $A$ aromaticity labels.   The model therefore learns
a direction-conditioned map
\begin{equation}
G_{\mathrm{source}}^d
\longrightarrow
\widehat G_{\mathrm{target}}^d,
\qquad
d\in\{\mathrm{forward},\mathrm{retro}\}.
\end{equation}
Rather than reconstructing the target graph entirely from scratch, UniRxnAxis parameterizes this map as a source-conditioned redistribution of graph structure and atom states.
For bond prediction, UniRxnAxis uses a residual structural form:
\begin{equation}
\widehat B_{\mathrm{target}}^d
= B_{\mathrm{source}}^d + \Delta \widehat B^d,
\end{equation}
while atom-level target properties are predicted from the decoder
representation. This mirrors the chemical fact that many reactions preserve a
large scaffold and concentrate the difficult prediction on sparse bond and
state changes.

\subsection{UniRxnAxis  Framework} 
\begin{figure}[t]
\centering
\includegraphics[width=0.9\textwidth]{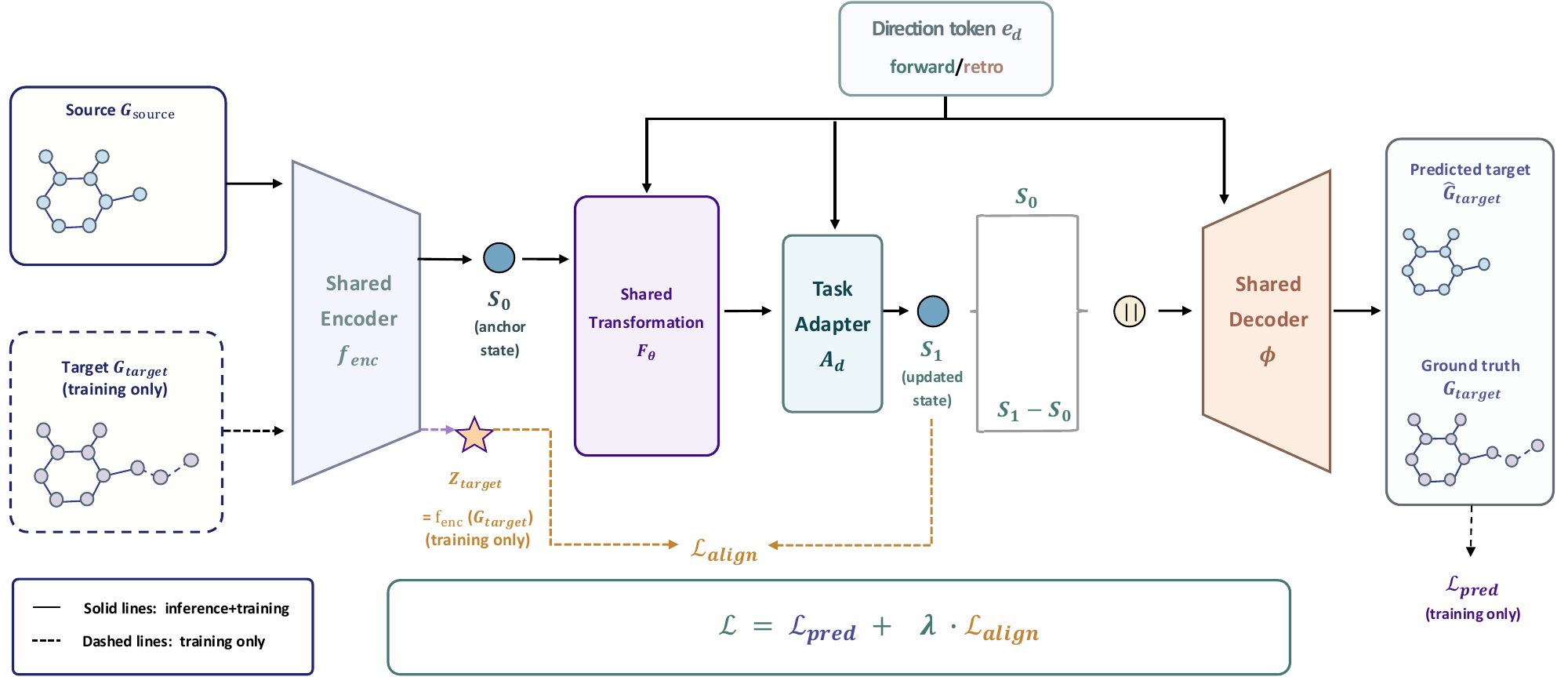} \vspace{-0.1in}
\caption{Overview of the UniRxnAxis architecture. It jointly models forward
prediction and retrosynthesis with shared endpoint encoders, a shared
direction-conditioned attention-style update module $F_\theta$, a
direction-conditioned adapter $A_d$, and a source-conditioned residual decoder. The
encoded target endpoint supplies the latent alignment reference used during
training and audit.}
\label{fig:urf_overview}
\end{figure}

\vspace{-0.1in}
UniRxnAxis consists of three coupled components, as shown in Fig. \ref{fig:urf_overview}. First, a shared molecular encoder $f_{Enc}$ maps each discrete reaction endpoint into a continuous  latent state. Second, a shared direction-conditioned transformation moves the encoded source toward the encoded target endpoint. In our implementation, this transformation is factorized into a module $F_\theta$ followed by a lightweight direction-conditioned adapter $A_d$. Third, a source-conditioned residual decoder $\phi$ combines the original source encoding with the predicted latent displacement to reconstruct the target graph.

In short, the pipeline is $G_{\mathrm{source}}\xrightarrow{f_{\mathrm{Enc}}}S_0\xrightarrow{A_d\circ F_\theta}S_1\xrightarrow{\phi([S_0\|S_1-S_0])}\widehat G_{\mathrm{target}}$, and $z_{\mathrm{target}}=f_{\mathrm{Enc}}(G_{\mathrm{target}})$ is used only as a training and audit reference, never as an inference-time input.
The same architecture is used for both forward prediction and retrosynthesis. Rather than training separate models, UniRxnAxis conditions the $F_\theta$, $A_d$, and decoder $\phi$ on a learned direction embedding $e_d$, where $d\in\{\mathrm{forward},\mathrm{retro}\}$. Thus, forward and retrosynthetic examples share the same encoder and transformation machinery, while the reaction direction is specified through learned conditioning.


\noindent
\textbf{Component 1: shared endpoint encoder.}
Given a source endpoint graph $G_{\mathrm{source}}^d$, the shared molecular
encoder produces a latent embedding
\begin{equation}
     S_0^d = f_{\mathrm{Enc}}(G_{\mathrm{source}}^d).
\end{equation}
The same encoder also encodes the target endpoint during training and audit:
\begin{equation}
z_{\mathrm{target}}^d
= f_{\mathrm{Enc}}(G_{\mathrm{target}}^d).
\end{equation}
The source and target endpoints are therefore represented by the same encoder
and occupy the same model-native latent coordinate system.



\noindent
\textbf{Component 2: direction-conditioned latent transformation.}
The direction-conditioned transformation is implemented as
\begin{equation}
S_0^d=H_0^d
\xrightarrow{\;F_\theta(\cdot;e_d)\;}
\widetilde H_1^d
\xrightarrow{\;A_d(\cdot;e_d)\;}
S_1^d,
\end{equation}
where $F_\theta$ produces an intermediate updated state and $A_d$ maps it to the
final direction-adapted state $S_1^d$,  which is trained to align with the
encoded target endpoint $z_{\mathrm{target}}^d$.
 

\noindent
\textbf{Component 3: source-conditioned residual decoding.}
The decoder reconstructs the target graph as:
\begin{equation}
\widehat G_{\mathrm{target}}^d
=
\mathcal  \phi\!\left(Z_{\mathrm{dec}}^d; G_{\mathrm{source}}^d, e_d\right),
\end{equation}
where  $ Z_{\mathrm{dec}}^d = f_{\mathrm{fuse}}\!\left( [\,S_0^d \mathbin{\|} (S_1^d-S_0^d)\,]; e_d \right)$.
Here, $\mathbin{\|}$ denotes feature concatenation, so
$Z_{\mathrm{dec}}^d$ fuses the original source state $S_0^d$ with the predicted
latent displacement $S_1^d-S_0^d$, 
conditioned on the direction embedding $e_d$.
In implementation, the bond decoder predicts residual edits relative to the
source graph rather than absolute target bonds from scratch. Let
$B_{\mathrm{source}}^d$ denote the source bond-order matrix extracted from
$G_{\mathrm{source}}^d$. Then
\begin{equation}
\widehat B_{\mathrm{target}}^d
=
B_{\mathrm{source}}^d
+
\mathcal \phi_{\mathrm{bond}}\!\left(Z_{\mathrm{dec}}^d; e_d\right).
\end{equation}
Additional decoder heads predict the remaining target-side atom states, such as
formal charges, aromaticity labels, and, in retrosynthesis, atom types for dummy
slots.  
  

\subsection{Training with Dual Endpoint Constraints } \vspace{-0.1in}
UniRxnAxis is trained with two coupled constraints: structured endpoint prediction and
latent endpoint alignment.  
First, for each direction $d$,  the structured endpoint prediction minimizes: 
\begin{equation}
\mathcal L_{\mathrm{pred}}^d
=
\mathcal L_{\mathrm{bond}}^d
+ \mathcal L_{\mathrm{charge}}^d
+ \mathcal L_{\mathrm{aroma}}^d
+ \mathbf 1[d=\mathrm{retro}]\,\mathcal L_{\mathrm{elem}}^d .
\end{equation}
Here, $\mathcal L_{\mathrm{bond}}^d$, $\mathcal L_{\mathrm{charge}}^d$, and
$\mathcal L_{\mathrm{aroma}}^d$ supervise bond orders, formal charges, and
aromaticity labels, respectively. The  
$\mathcal L_{\mathrm{elem}}^d$  is  used only in retrosynthesis, where product-to-reactant
decoding must fill dummy atom slots on the reconstructed reactant side.   Each term is implemented as a masked cross-entropy loss over its corresponding discrete target labels.


Second, UniRxnAxis aligns the transformed source state with the encoded target
endpoint. During training, the target graph is encoded as
$z_{\mathrm{target}}^d$, and $S_1^d$ is encouraged to approach it:
\begin{equation}
\label{eq:align_loss}
\mathcal L_{\mathrm{align}}^d
=
\frac{
\sum_i m_i^d
\left\|S_{1,i}^d-z_{\mathrm{target},i}^d\right\|_2^2
}{
\sum_i m_i^d
},
\end{equation}
where $m_i^d$ selects valid target/active slots. The full objective is
\begin{equation}
\mathcal L
=
\mathbb E_d
\left[
\mathcal L_{\mathrm{pred}}^d
+ \lambda_{\mathrm{align}}\mathcal L_{\mathrm{align}}^d
\right],
\end{equation}
where $\lambda_{\mathrm{align}}$ controls the strength of the latent endpoint-alignment constraint. 

It is worth noting that the latent alignment constraint specifies where the target endpoint lies in representation space, but does not prescribe how the direction-conditioned stack $(F_\theta,A_d,\phi)$ transforms $S_0$ into the decoded target. Whether the learned update follows the direct endpoint direction, and whether deviations from that direction are functionally important, therefore remains an empirical question addressed by the directional audit introduced next.


\vspace{-0.1in}
\subsection{Directional Decomposition} \vspace{-0.1in}
The dual constraints make directional auditing possible: the prediction loss trains the model to reconstruct the target endpoint, while the alignment loss defines a target direction in latent space against which the learned update can be decomposed.

For each reaction and direction, the dual constraints define two latent
displacements. The supervision-implied endpoint displacement is
\begin{equation}
g^d = z_{\mathrm{target}}^d - S_0^d,
\end{equation}
while the model's realized update is
\begin{equation}
\Delta z^d = S_1^d - S_0^d .
\end{equation}
Let $\hat g^d=g^d/\|g^d\|$. We decompose $\Delta z^d$ into an
endpoint-parallel component and an endpoint-orthogonal residual:
\begin{equation}
\Delta z_{\parallel}^d
=
\langle \Delta z^d,\hat g^d\rangle \hat g^d,
\qquad
\Delta z_{\perp}^d
=
\Delta z^d-\Delta z_{\parallel}^d .
\end{equation}
The endpoint-parallel counterfactual state is then
\begin{equation}
S_{\mathrm{mid}}^d = S_0^d+\Delta z_{\parallel}^d .
\end{equation}

We interpret $\Delta z_{\parallel}^d$ as \textbf{\emph{endpoint-aligned attraction}}: the
part of the learned update that points toward the encoded target endpoint. In
contrast, we call $\Delta z_{\perp}^d$ \textbf{\emph{emergent steering}}: the
endpoint-orthogonal part of the update that is produced by the model, is not
directly specified by the alignment loss, but may still influence decoder
decisions.
The decomposition is relative to the reference axis $g^d$, which need not be decoder-executable and requires the ground-truth endpoint; $\Delta z_{\perp}^d$ is thus an audit-time quantity (see Sec.~\ref{sec:axis_dependence}).
This interpretation is falsifiable. If $\Delta z_{\perp}^d$ is only noise,
random replacements or shuffles should preserve decoding. If it carries a real
steering signal, then removing, replacing, or scaling it should change decoder
decisions, and its mass should align with reaction-relevant structure.  Section~4 therefore tests whether \emph{emergent steering}
$\Delta z_{\perp}^d$ is functionally necessary, chemically organized, and
distinct from random or shuffled alternatives.

\vspace{-0.1in}
\section{Experiments}
\vspace{-0.1in}

\paragraph{Metric conventions.}
In this section, all audits use USPTO-50K ~\cite{schneider2016s} data following standard splits.
Code is available at \url{https://github.com/yilish/UniRxnAxis}.
Reaction classification means predicting the USPTO-50K reaction-class label (10 classes)
from a frozen latent representation with a lightweight linear probe. A bond
edit is any atom-pair bond whose order changes between the source and target
molecules; formed bonds are the subset absent in the source and present in the
target, while broken bonds are present in the source and absent in the target.
Edit, formed, broken, and center F1 compare decoded predictions against these
bond-edit tensors, with center atoms defined as atoms incident to any
bond edit. Bond-logit probes use the model's continuous bond logits before
 rounding into  bond orders.

We organize the evaluation around six questions.

\textbf{Q1: Is the audited model a valid substrate?}
Before interpreting internal latent directions, the model must first be an
effective endpoint predictor for both forward prediction and retrosynthesis.

\textbf{Q2: Does endpoint alignment explain the executable update?}
After evaluating decoding on \(S_1\) (in Q1), we decode the
endpoint-parallel state
\(
S_{\mathrm{mid}} = S_0+\Delta z_{\parallel},
\)
 and  the residual-only state \(S_0+\Delta z_{\perp}\). If
\(S_{\mathrm{mid}}\) performs poorly while the residual-only state recovers most
of the full \(S_1\) performance, endpoint attraction is not a sufficient
decoder-executable explanation of the learned update.

\textbf{Q3: Is the residual structured and sample-specific rather than an artifact?}
To test whether the residual's utility is an artifact of neural networks, we
replace \(\Delta z_{\perp}\) with norm-matched random directions, token/sample
shuffles, same-class shuffles, and class prototypes.

\textbf{Q4: Does the residual act through specific decoder bond decisions?}
We inspect bond logits before final discretization,
including boundary crossing, target-error reduction, and local
\(d\,\mathrm{Bond}/dz\) sensitivity. This tests whether \(\Delta z_{\perp}\)
acts on specific bond edits rather than merely moving the latent state into a
globally decoder-preferred region.

\textbf{Q5: Is residual steering organized around reaction-relevant structure?}
We test whether update components and residual-heavy atoms are related to
reaction type, reaction-center enrichment, and atom-level suppress/rescue
effects.

\textbf{Q6: Can the decomposition support model-side failure diagnosis?}
Finally, we use the decomposition to build a proxy taxonomy that separates
endpoint symptoms from latent-stage diagnostic hypotheses. As a narrow proof of
concept, we also test whether residual dosage exposes local correction headroom
for a subset of near-miss failures.



\vspace{-0.1in}
\subsection{Q1: Is the audited model a valid substrate?} \vspace{-0.1in}

Table~\ref{tab:endpoint_baseline_context} reports the benchmarking result under the standard USPTO-50K retrosynthesis, and select representative baselines across template-based, semi-template-based, template-free, unified reaction-model, and large-pretraining paradigms.  The joint UniRxnAxis checkpoint is competitive on USPTO-50K, 
reaching $64.4\%$ retrosynthesis.  On forward prediction tasks, our model reaches Top-1 accuracy and $88.3\%$ under joint training and $88.7\%$ under forward only training settings. 

\begin{table}[h!]
\centering
\begin{minipage}[b]{0.50\linewidth}
\centering
\scriptsize
\setlength{\tabcolsep}{3pt}
\caption{USPTO-50K retrosynthesis top-1 accuracy with reaction class unknown.}
\label{tab:endpoint_baseline_context}
\resizebox{\linewidth}{!}{%
\begin{tabular}{llr}
\toprule
\textbf{Paradigm} & \textbf{Model} & \textbf{Top-1} \\
\midrule
\multirow{2}{*}{Template-based}
& GLN ~\citep{dai2019gln} & 52.5 \\
& LocalRetro ~\citep{chen2021localretro} & 53.4 \\
\midrule
\multirow{2}{*}{Semi-template-based}
& RetroPrime ~\citep{wang2021retroprime}& 51.4 \\
& Graph2Edits ~\citep{zhong2023graph2edits} & 55.1 \\
\midrule
\multirow{3}{*}{Template-free}
& Graph2SMILES ~\citep{tu2022permutation} & 52.9 \\
& NAG2G ~\citep{yao2024node} & 55.1 \\
& EditRetro ~\citep{han2024iterative}& 60.8 \\
\midrule
\multirow{2}{*}{Unified}
& T5Chem ~\citep{lu2022unified}& 46.5 \\
& RXNGraphormer ~\citep{xu2025unified} & 51.0 \\
\midrule
Pretraining
& RSGPT ~\citep{deng2025rsgpt} & 63.4 \\
\midrule
\multirow{2}{*}{Ours}
& UniRxnAxis (retro only) & 61.7 \\
& \textbf{UniRxnAxis (joint)} & \textbf{64.4} \\
\bottomrule
\end{tabular}}
\end{minipage}\hfill
\begin{minipage}[b]{0.46\linewidth}
\centering
\scriptsize
\setlength{\tabcolsep}{3pt}
\caption{Audit substrate checks: joint training improves endpoint role readout (top)
and native forward--retro update retrieval (bottom).}
\label{tab:audit_substrate}
\label{tab:audit_role_readout}
\label{tab:audit_update_alignment}
\resizebox{\linewidth}{!}{%
\begin{tabular}{llrr}
\toprule
\multicolumn{4}{l}{\textit{Reaction-type prediction from $S_1$ (macro-F1)}} \\
\midrule
Endpoint role & Forward-only & Joint & Retro-only \\
\midrule
Product as $S_1$ / forward target & 0.4658 & \textbf{0.7731} & 0.7299 \\
Product as $S_0$ / retro source & 0.5674 & \textbf{0.7257} & 0.6846 \\
Reactant as $S_0$ / forward source & 0.7103 & \textbf{0.8243} & 0.6842 \\
Reactant as $S_1$ / retro target & 0.6950 & \textbf{0.8172} & 0.6532 \\
\midrule
\multicolumn{4}{l}{\textit{Forward--retro matching retrieval}} \\
\midrule
Setting & Object & AUC & R@1 \\
\midrule
Split direct & $\Delta z_{\perp}$ & 0.5131 & 0.0000 \\
Split-aligned & $S_0$ & 0.9404 & \textbf{0.5574} \\
Split-aligned & $S_1$ & 0.9437 & 0.4068 \\
Split-aligned & $\Delta z_{\perp}$ & 0.9472 & 0.3393 \\
Joint UniRxnAxis & $\Delta z_{\perp}$ & \textbf{0.9796} & 0.4907 \\
\bottomrule
\end{tabular}}
\end{minipage}
\end{table}

Directional audit also requires both reactant-side and product-side endpoints
to exist as meaningful latent states. Joint training of both forward and retro direction enables better semantic representation across all settings. Reaction-class classification proves this pattern 
 across all tasks (Table~\ref{tab:audit_substrate}, top). 

Bidirectional training also enables a more aligned transformation geometry. We test the forward--retro retrieval, which treats one
direction's latent object as a retrieval query and asks whether the paired
opposite-direction object is nearest among test examples. AUC and R@1 therefore
measure whether forward/retro transformation objects are directly comparable in
the same coordinate system.
 Without alignment, split forward/retro checkpoints leave update objects near chance. 
 Post-hoc split-space alignment can recover substantial similarity, but it requires an extra learned map; joint UniRxnAxis exposes $\Delta z_{\perp}$ comparability directly in native coordinates (Table~\ref{tab:audit_substrate}, bottom). Thus bidirectional training appears to organize the decomposed transformation component in a shared audit space rather than only making static endpoint states alignable.


\vspace{-0.15in}
\subsection{Q2: Does endpoint alignment explain the executable update?}

\vspace{-0.1in}
\begin{wraptable}{r}{0.45\linewidth}
\centering
\small
\setlength{\tabcolsep}{3pt} \vspace{-0.1in}
\caption{Per-sample Geometry metrics.}
\label{tab:endpoint_geometry}
\resizebox{0.98\linewidth}{!}{
\begin{tabular}{lrr}
\toprule
Dir. & Mean $\cos(g, \Delta z)$ & Mean $\|\Delta z_{\perp,i}\|/\|\Delta z_i\|$\\
\midrule
Forward & 0.3282 & 0.9322 \\
Retro & 0.6273 & 0.7405 \\
\bottomrule
\end{tabular}}\vspace{-0.1in}
\end{wraptable}
Minimizing (eq. ~\ref{eq:align_loss}) will move the model update $\Delta z$  toward 
the direct endpoint direction $g=z_{\mathrm{target}}-S_0$. 
Table~\ref{tab:endpoint_geometry} reports a stronger relevance of  $\Delta z$ and $g$ in retrosynthesis than in  forward prediction.  
However, the large orthogonal fraction (mean $r_{\perp}=0.74$)
shows that much of the update lies outside the endpoint direction. The target
encoding thus acts mainly as a reference axis around which we measure the learned motion.

Geometry alone does not tell us whether the endpoint-parallel motion is useful to the decoder. 
We therefore decode the endpoint-parallel state
$S_{\mathrm{mid}} = S_0+\Delta z_{\parallel}$.
This intervention keeps only the part of the learned update explained by the target-reference direction and removes the endpoint-orthogonal residual \(\Delta z_{\perp}\). 
If endpoint alignment were sufficient for the transformation, \(S_{\mathrm{mid}}\) should retain most of the full decoding performance.

In retrosynthesis, however, this is not the case. 
Table~\ref{tab:retro_stage_metrics} shows that the endpoint-parallel state reaches only \(0.0997\) top-1 accuracy, far below the full transformed state \(S_1\) at \(0.6441\). 
Endpoint-directed motion alone is thus not a sufficient decoder-executable retro program; the decoder needs update content beyond \(\Delta z_{\parallel}\).
\vspace{-0.1in}
\begin{table}[h!]
\centering
\begin{minipage}[b]{0.45\linewidth}
\centering
\scriptsize
\setlength{\tabcolsep}{3pt}
\caption{Retro latent-state decoding. Residual-only (\(S_0+\Delta z_{\perp}\)) nearly matches full
$S_1$, while $S_{\mathrm{mid}}$ and $z_\text{target}$ remain weak.}
\label{tab:staged_decoding}
\label{tab:functional_split}
\label{tab:retro_stage_metrics}
\resizebox{\linewidth}{!}{%
\begin{tabular}{lrrrr}
\toprule
State & Top-1 & Edit & Formed & Center \\
\midrule
$S_0$ & 0.0405 & 0.3832 & 0.0684 & 0.4674 \\
$S_{\mathrm{mid}}$ & 0.0997 & 0.4998 & 0.1927 & 0.6046 \\
\(S_0+\Delta z_{\perp}\) & \textbf{0.6303} & 0.8799 & 0.8761 & 0.9371 \\
$S_1$ & \textbf{0.6441} & \textbf{0.8868} & \textbf{0.8868} & \textbf{0.9418} \\
$z_{\mathrm{target}}$ & 0.1110 & 0.5082 & 0.2071 & 0.6138 \\
\bottomrule
\end{tabular}} 
\end{minipage}\hfill 
\begin{minipage}[b]{0.52\linewidth}
\centering
\scriptsize
\setlength{\tabcolsep}{3pt} 
\caption{$\Delta z_\perp$ artifact controls. The true $\Delta z_\perp$ works; norm,
prototype, and shuffle controls fail.} \vspace{-0.1in}
\label{tab:artifact_controls}
\label{tab:norm_controls}
\label{tab:prototype_shuffle_controls}
\resizebox{0.5\linewidth}{!}{%
\begin{tabular}{llr}
\toprule
Control & Variant & Top-1 \\
\midrule
\multirow{2}{*}{Reference} & Full $S_1$ & \textbf{0.6441} \\
 & $S_0+\Delta z_\perp$ & \textbf{0.6303} \\
\midrule
\multirow{2}{*}{Norm} & Global & 0.1007 \\
 & Per-token & 0.1022 \\
\midrule
\multirow{3}{*}{Prototype} & Class subspace & 0.2606 \\
 & Class NN & 0.1460 \\
 & Class mean & 0.1558 \\
\midrule
\multirow{3}{*}{Shuffle} & Token & 0.0855 \\
 & Sample & 0.1162 \\
 & Same-class & 0.1188 \\
\bottomrule
\end{tabular}}
\end{minipage}
\end{table} \vspace{-0.15in}

The audit geometry is bidirectional, but the two directions are not equally
diagnostic. Forward prediction is empirically easier, with a more compact
$S_1$ geometry (Appendix~\ref{app:effective_dimension}). The high forward $S_0$ baseline is also
consistent with the copy-scaffold inductive bias of the residual decoder being
nearly sufficient for forward prediction, making forward a negative-control
direction for the audit rather than the main analysis target. In contrast, retrosynthesis is a harder one-to-many inverse
problem with disconnection ambiguity, and therefore provides a more sensitive
setting for auditing whether the $\Delta z_\perp$ carries
decoder-relevant signal. We therefore focus the  analyses mainly on
the retro direction.
Morever,  $\Delta z_\perp$ is dominant in retrosynthesis, but $\Delta z_{\parallel}$ is
not merely decorative: when the true residual is fixed, the full state still
outperforms residual-only, while anti-parallel or shuffled parallel vectors
degrade performance (Appendix Table~\ref{tab:parallel_preconditioner}).

\vspace{-0.1in}
\subsection{Q2 (cont.): Is the asymmetry an artifact of the reference axis or the model?}
\label{sec:axis_dependence}
\vspace{-0.1in}
The encoded target \(z_{\mathrm{target}}\) itself decodes at only \(0.1110\) (Table~\ref{tab:staged_decoding}), and the decoder is trained only on realized displacements \(S_1-S_0\). The weak \(S_{\mathrm{mid}}\) could therefore reflect a reference axis that the decoder cannot execute, rather than a property of the learned update. We test this alternative in four ways; details are in Appendix~\ref{app:rebuttal_analyses}.

\begin{wraptable}{r}{0.52\linewidth}
\centering
\scriptsize
\setlength{\tabcolsep}{2.5pt}\vspace{-0.15in}
\caption{Retro sensitivity to \(\lambda_{\mathrm{align}}\) (test top-1, \%). \(r_\perp=\mathbb E\|\Delta z_\perp\|/\|\Delta z\|\). \(0.1\) is the audited checkpoint.}
\label{tab:lambda_sweep}
\begin{tabular}{lrrrrrrr}
\toprule
\(\lambda_{\mathrm{align}}\) & \(r_\perp\) & cos & Par. & Res. & \(S_1\) & \(z_{\mathrm{t}}\) & Res.\(-\)Par. \\
\midrule
0.001 & .738 & .649 & 18.41 & 55.74 & 60.56 & 14.88 & +37.33 \\
0.01 & .649 & .734 & 49.69 & 57.96 & 59.62 & 49.39 & +8.27 \\
0.1 & .741 & .627 & 9.97 & 63.03 & 64.41 & 11.10 & +53.06 \\
1 & .414 & .869 & 23.21 & 39.92 & 52.99 & 24.97 & +16.71 \\
10 & .157 & .934 & 11.10 & 28.28 & 49.95 & 11.10 & +17.18 \\
\bottomrule
\end{tabular}\vspace{-0.15in}
\end{wraptable}
\textbf{Alignment strength.} We retrain with \(\lambda_{\mathrm{align}}\in\{0.001,0.01,1,10\}\) under the same budget (Table~\ref{tab:lambda_sweep}). Geometry changes monotonically: cosine alignment rises and the residual fraction falls as \(\lambda_{\mathrm{align}}\) grows. At \(\lambda_{\mathrm{align}}=0.01\) the target axis becomes substantially executable (\(z_{\mathrm{t}}=49.39\), Parallel \(=49.69\)), and the gap shrinks to \(+8.27\) points, yet Residual remains higher. At \(\lambda_{\mathrm{align}}=10\) the residual is only \(15.7\%\) of the update norm, yet it still exceeds Parallel by \(17.18\) points, at the cost of lower \(S_1\) accuracy.

\textbf{Decoder-interface calibration.} With the model frozen, a label-free, zero-intercept linear map \(T(g)=gW\), fit on the training split to match realized displacements, raises axis decoding from \(11.10\%\) to \(21.25\%\). Re-decomposing around \(T(g)\) still gives \(52.73\%\) Residual versus \(21.25\%\) Parallel. Fine-tuning the decoder on \([S_0\|g]\) raises \(z_{\mathrm{target}}\) decoding to \(64.45\%\), but also yields \(10.39\%\) with the displacement zeroed. This reveals an \(S_0\) shortcut, so we read it as a supervised-recoverability upper bound rather than pre-existing axis utility.

\textbf{Matched-random axes.} For each test reaction we sample eight null axes with the same angle to \(\Delta z\) as \(g\), which matches component norms and only changes orientation. The target-defined axis gives a \(53.07\)-point Residual--Parallel separation versus \(37.13\) for matched axes (difference \(15.94\), 95\% CI \([14.83,17.10]\)). Generic projection geometry thus contributes but does not explain the full effect. Scale mismatch is not dominant either (\(\mathbb E\|g\|/\|\Delta z\|=0.95\); decoding \(S_0+c\,g\) peaks at \(19.53\%\)).

\textbf{Other architectures and datasets.} We add an endpoint-alignment loss to the autoregressive graph-edit model Graph2Edits~\citep{zhong2023graph2edits} (\(55.44\%\) top-1) and audit the state change across the ground-truth first edit. Its frozen edit head predicts the second edit at \(16.58\%\) from the parallel-only state versus \(48.79\%\) from the residual-only state (\(S_1\): \(75.80\%\)). On USPTO-MIT~\citep{jin2017predicting}, retrosynthesis again remains more endpoint-aligned than forward prediction (\(\Delta\cos\approx+0.295\), \(\Delta r_\perp\approx-0.17\)). This transfer is geometric only; we do not test decoder executability there.

\textbf{Interpretation.} The gap's magnitude depends on the reference axis, the alignment weight, and the decoder interface; its sign is stable in every setting we tested. Two factors contribute. The decoder is exposed only to realized updates, which explains part of the gap. In addition, \(\mathcal L_{\mathrm{align}}\) only softly attracts \(S_1\) toward the endpoint, while \(\mathcal L_{\mathrm{pred}}\) demands exact, sample-specific bond decisions that can occupy degrees of freedom endpoint attraction leaves free. Our claim is therefore not that orthogonal directions are intrinsically more useful, but that, relative to the supervised endpoint axis, the native decoder-executable computation concentrates in the residual more than for matched arbitrary axes.



\vspace{-0.1in}
\subsection{\texorpdfstring{Q3: Is $\Delta z_\perp$ structured and sample-specific rather than an artifact?}{Q3: Is Delta z-perp structured and sample-specific rather than an artifact?}}
\vspace{-0.1in}
\textbf{$\Delta z_\perp$ is not a magnitude-only artifact.}
A functional residual could still be an artifact if its utility came from
magnitude alone. Norm-matched random controls do  not recover the true
residual's performance or edit-boundary crossing (Table~\ref{tab:artifact_controls}, top), which rejects this hypothesis.

\textbf{$\Delta z_\perp$ is not template or shuffle shortcuts.}
A second hypothesis is that the residual may be a class-level template direction
rather than a sample-specific steering signal. Class means, nearest-neighbor
class residuals, and class subspace projections fall far below the true
residual and the full model (Table~\ref{tab:artifact_controls}, bottom). Nor
is any nearby empirical residual good enough: token shuffling, sample shuffling,
and same-class shuffling all collapse accuracy. Thus, the residual is therefore not a
norm, prototype, or shuffle-stable artifact. Instead, it is sample-specific.

\vspace{-0.1in}
\subsection{\texorpdfstring{Q4: Does $\Delta z_\perp$ act through specific decoder bond decisions?}{Q4: Does Delta z-perp act through specific decoder bond decisions?}}
\vspace{-0.1in}

The previous controls show that the residual is structured, but they do
not yet show where it acts in the decoder. A remaining hypothesis is that \textbf{$\Delta z_\perp$ merely moves the latent state into a globally
decoder-preferred region, rather than pushing bond-edit
decisions.} We test this hypothesis from three decoder-facing views: boundary
crossing, sensitivity, bond logit refinements.

\textbf{$\Delta z_\perp$ selectively pushes bonds across decoder boundaries.}
We compare bond logits before and after adding the residual step from
$S_{\mathrm{mid}}$ to $S_1$. As shown in
Table~\ref{tab:bond_boundary_derivative}A, $\Delta z_\perp$ selectively pushes
correct edit and formed edges across decoding thresholds, rarely pushes wrong
edges across thresholds, and suppresses competing wrong edges. We quantify this
using target-edge crossing, wrong-edge crossing, and competing-edge suppression.
The high target-edge crossing and near-zero wrong-edge crossing indicate that
emergent steering acts selectively on bond-edit decisions rather than globally
increasing decoder confidence.

\textbf{Decoder sensitivity is strongest along $\Delta z_\perp$.}
The decoder's local score
sensitivity is concentrated along $\Delta z_{\perp}$ on the same edges where
the residual improves finite decoding behavior. Finite interventions are
complemented by a local derivative test. Among tested
directions, $\Delta z_{\perp}$ has the strongest local directional derivative
on target formed bonds and the strongest negative derivative on selected
competing non-target edges, much larger than $\Delta z_{\parallel}$ or a
random direction (Table~\ref{tab:bond_boundary_derivative}, Panel B). 

\textbf{\texorpdfstring{$\Delta z_\perp$}{Delta z-perp} selectively refines bond logits.}
We next test whether the boundary effects reflect targeted score correction or a broad score shift. Using pre-discretization bond logits, we measure bond-logit shift and target-error reduction (TER) for target edit/formed bonds and high-scoring false-positive competitors. The $\Delta z_\parallel$ step gives coarse support by raising target-edge scores, but also increases false-positive competitors. In contrast, $\Delta z_\perp$ raises target edit/formed-edge scores more strongly while suppressing competitors, yielding the largest TER on target edges and positive TER on false positives (Table~\ref{tab:score_delta_ter}C). Thus, $\Delta z_\perp$ refines bond logits toward the target pattern rather than simply amplifying decoder confidence.


\begin{table}[t]
\centering
\scriptsize
\setlength{\tabcolsep}{3pt}
\caption{Decoder-facing bond probes. $\Delta z_{\perp}$ crosses target
thresholds, has the strongest tested local derivative, and selectively refines
bond logits.} \vspace{-0.05in}
\label{tab:bond_boundary_derivative}
\label{tab:boundary_crossing}
\label{tab:directional_derivative}
\label{tab:score_delta_ter}
\begin{minipage}[t]{0.29\linewidth}
\centering
\resizebox{\linewidth}{!}{%
\begin{tabular}{lr}
\toprule
\multicolumn{2}{l}{\textit{A: boundary crossing}} \\
\midrule
Target edit crosses & \textbf{0.5217} \\
Target formed crosses & \textbf{0.6054} \\
Wrong non-target crosses & \textbf{0.0004} \\
Wrong competitors down & \textbf{0.8525} \\
\bottomrule
\end{tabular}}
\end{minipage}\hfill
\begin{minipage}[t]{0.29\linewidth}
\centering
\resizebox{\linewidth}{!}{%
\begin{tabular}{lr}
\toprule
\multicolumn{2}{l}{\textit{B: local derivative}} \\
\midrule
$\Delta z_{\perp}$ on target formed & \textbf{+1.0401} \\
$\Delta z_{\parallel}$ on target formed & +0.1839 \\
Random on target formed & -0.0017 \\
$\Delta z_{\perp}$ on competitors & \textbf{-0.4104} \\
\bottomrule
\end{tabular}}
\end{minipage}\hfill
\begin{minipage}[t]{0.39\linewidth}
\centering
\resizebox{\linewidth}{!}{%
\begin{tabular}{lrrrr}
\toprule
\multicolumn{5}{l}{\textit{C: bond-logit refinement}} \\
\midrule
Edge group & \multicolumn{2}{c}{Shift} & \multicolumn{2}{c}{TER} \\
\cmidrule(lr){2-3}\cmidrule(lr){4-5}
 & $\parallel$ & $\perp$ & $\parallel$ & $\perp$ \\
\midrule
Target edit & 0.1926 & \textbf{0.3740} & 0.1611 & \textbf{0.5143} \\
Target formed & 0.3288 & \textbf{0.5129} & 0.1425 & \textbf{0.6721} \\
Wrong competitors & 0.5532 & \textbf{-0.5427} & -0.5542 & \textbf{0.5423} \\
\bottomrule \vspace{-0.3in}
\end{tabular}}
\end{minipage}
\end{table}

A scalar residual-dosage sweep also
peaks near the learned residual magnitude rather than growing monotonically 
(Appendix Table~\ref{tab:retro_dosage}).
\vspace{-0.1in}
\subsection{Q5: Is residual steering organized around reaction-relevant structure?}
\vspace{-0.1in}

If the residual is part of reaction computation, it should reflect reaction-level organization, not only endpoint accuracy. Stage-wise probes show that reaction-class readability increases from early encoder layers to \(S_0\), then decreases after the \(S_0\!\rightarrow\!S_1\) update. In retrosynthesis, however, the decomposed components \(\Delta z_{\parallel}\) and \(\Delta z_{\perp}\) are more class-readable than the raw update \(\Delta z\) (Appendix Figure~\ref{fig:stage_class_readout}). This suggests a program-to-refinement pattern: the encoder builds a class-readable reaction program, while the update converts it into an instance-specific bond-edit state for decoding.
Appendix~\ref{app:delta_space_viz} includes a qualitative joint
forward/retro t-SNE visualization of the same update components, used only as
supporting semantic visualization rather than primary evidence.

As supporting chemistry-facing evidence, atom-level diagnostics show
that update components are enriched around reaction centers. Under the
raw-anchor convention, $\Delta H_{\parallel}$ carries the largest retro
reaction-center mass ($0.4918$), while $\Delta H_{\perp}$ remains enriched over the label
fraction ($0.3255$ vs.\ $0.1596$). The residual is thus not the most localized
component: $\Delta z_{\parallel}$ already points to the reaction center, while
$\Delta z_{\perp}$ selects among nearby candidate edits (Table~\ref{tab:bond_boundary_derivative}). Residual-heavy atoms are also functionally
tied to decoding: suppressing top residual atoms strongly degrades formed-bond
F1, and rescuing them recovers part of the loss
(Appendix~\ref{app:atom_sample_diagnostics}). These supporting
evidence suggests that residual structure is chemistry-facing.


\vspace{-0.1in}
\subsection{Q6: Can the decomposition support model-side failure diagnosis?}
\vspace{-0.1in}

\textbf{UniRxnAxis provides two-level failure labels.}
Separating $\Delta z_\parallel$ from $\Delta z_\perp$ makes failure diagnosis a
two-level question: what is wrong in the decoded endpoint, and where along
$S_0\!\rightarrow\!S_{\mathrm{mid}}\!\rightarrow\!S_1$ does the transformation
become unreliable? We therefore build a
hierarchical, proxy-based taxonomy with endpoint symptoms on the output side
and latent-stage diagnostic hypotheses on the model side. We present the taxonomy in Appendix~\ref{app:failure_taxonomy_definitions} 
in detail.

Both levels use five matched categories: program (parallel), refinement
(residual), dosage, nonresponsive/off-manifold, and decoder-realization. Among
$1782$ wrong retro test samples, the merged endpoint taxonomy assigns a
non-mixed label to all samples, and the latent taxonomy assigns a non-mixed
label to $96.6\%$ (Appendix Figure~\ref{fig:merged_failure_taxonomy}A--B).

\textbf{ Off-diagonal audit provides insight into models' internal behavior.}
The endpoint--latent crosstab first acts as a sanity check: the
conceptual diagonal carries substantial mass (mean available row fraction
$0.650$), so the taxonomy recovers obvious stage correspondences when endpoint
symptoms and latent traces agree.
The most informative cases are off-diagonal: the endpoint symptom says
one thing, but the latent trajectory suggests another. The largest such pattern
is endpoint refinement/edge-selection errors mapping to latent
program/parallel failures ($503/1217$, row fraction $0.413$), rather than to
latent residual refinement failures ($433/1217$, row fraction $0.356$). Some apparent final competing-edge mistakes may thus originate in an
atypical or undercommitted program state before residual refinement.
Appendix Figure~\ref{fig:merged_failure_taxonomy}D shows one such case: the final output
has a local-looking edge/refinement symptom, while the decoded
$S_{\mathrm{mid}}$ state and proxy trace point upstream to program formation.



\textbf{On residual-dosage headroom.} 
The failure taxonomy suggests a broader implication: the residual is not
only a diagnostic signal, but also a place where the model may already contain
recoverable solutions. 
We choose one simple one-dimensional proof of concept: residual
dosage. \textbf{If some wrong samples fail mainly because the residual amount is
slightly under- or over-applied, can a different residual dosage recover them?}
We first measure the headroom with an oracle over the fine grid
$\alpha\in\{0,0.125,\ldots,2\}$ for
$S(\alpha)=S_{\mathrm{mid}}+\alpha\Delta z_{\perp}$. This oracle  identifies $247$
failures where $\alpha=1$ that could be corrected by changing only $\alpha$, raising
official accuracy from $0.6441$ to $0.6934$. We then test a  simple
metric-gated selector: keep $\alpha=1$ by default, apply dosage changes only to
samples whose decoded proxy signals look dosage-sensitive, and choose the
alternative $\alpha$ from the fine-grid proxy score
(Appendix~\ref{app:adaptive_alpha_selector}). The best conservative rule
reaches $0.6529$, rescuing $46/247$ oracle-rescuable failures ($18.6\%$) while
breaking $2/3225$ alpha-$1$ correct samples ($0.06\%$)
(Table~\ref{tab:adaptive_alpha_rescue}).
\begin{table}[h]
\centering
\scriptsize
\setlength{\tabcolsep}{4pt}
\caption{Fine-grid residual-dosage rescue on the retro test split.}\vspace{-0.1in}
\label{tab:adaptive_alpha_rescue}
\begin{tabular}{lrrrrrr}
\toprule
Run & Fixed $\alpha=1$ & Oracle & Oracle-rec. & Selector & Rescue & Break \\
\midrule
Retuned fine grid & 0.6441 & 0.6934 & 247 & \textbf{0.6529} & 46 ($18.6\%$) & 2 ($0.06\%$) \\
\bottomrule \vspace{-0.3in}
\end{tabular}
\end{table}

The claim is not that scalar $\alpha$ solves retrosynthesis, but that the
audit exposes a narrow subset of near-miss failures whose decoded outcome
changes when the residual is dosed differently. Because $\Delta z_\perp$ itself
requires the ground-truth endpoint, this is not yet a deployable corrector.


\vspace{-0.1in}
\section{Conclusion} \vspace{-0.05in}

Endpoint correctness is necessary for reaction modeling, but it is not enough
to make a model scientifically inspectable. UniRxnAxis aligns both sides of the
reaction in a shared latent geometry, which enables decomposing the learned
update into endpoint-parallel attraction and endpoint-orthogonal
\emph{emergent steering}. Emergent steering preserves most retrosynthesis
utility, survives control substitutions, and modulates bond-edit decisions; its
advantage shrinks but persists when the reference axis is made more executable. Reaction models should thus be evaluated not only by
their endpoints but also by how their latent transformations organize chemical change.

\textbf{Limitations.} The decomposition uses the ground-truth endpoint, so
emergent steering is an audit-time diagnostic; the gap's size depends on the
reference axis, $\lambda_{\mathrm{align}}$, and the decoder interface, and only
its sign is stable in our tests. Executable evidence covers retrosynthesis on
USPTO-50K with two architectures, with geometric transfer only on USPTO-MIT,
and we report deterministic top-1 only. Reaction-center and failure labels are
unvalidated automatic proxies.


\begin{ack}
This work was supported by the National Science Foundation under the NSF
Center for Computer Assisted Synthesis (C-CAS), grant number CHE-2202693.

\textbf{Competing interests.} The authors declare no competing interests.
\end{ack}

\bibliographystyle{plainnat}
\bibliography{references}
\appendix
\section{Appendix}


\subsection{Audit State and Metric Definitions}
\label{app:audit_protocol}

Table~\ref{tab:audit_state_defs_app} summarizes the latent states used in the
counterfactual audit. The decomposition states depend on
$z_{\mathrm{target}}$, the encoded ground-truth endpoint latent, and are
therefore analysis-time states rather than deployable inference-time states.
The implemented endpoint chain is
$S_0=H_0=\mathrm{Enc}(G_{\mathrm{source}})
\xrightarrow{F_\theta(\cdot;e_d)}\widetilde H_1
\xrightarrow{A_d(\cdot;e_d)}S_1$. Thus $S_0$ is the raw encoded source, while
$S_1$ is the transformed source state aligned to
$z_{\mathrm{target}}=\mathrm{Enc}(G_{\mathrm{target}})$. The encoder is shared
and direction-agnostic; the update module, adapter, and decoder interface all
receive $e_d$.

Latent tensors are represented internally as token states
$X\in\mathbb R^{B\times L\times D}$. We use two aggregation domains. For
active-target graph-level readouts, let $m_{bi}=1$ when token $i$ is active in
the update window and valid on the target endpoint, matching the
endpoint-alignment mask
$m=(\neg\,\mathrm{target\_mask})\land\mathrm{active\_mask}$. The graph
readout is the masked mean
\[
\mathrm{Pool}_{m}(X)_b
=
\frac{\sum_i m_{bi}X_{bi}}{\max(1,\sum_i m_{bi})}\in\mathbb R^D .
\]
For encoder-layer stage probes that do not use a target endpoint, the same
masked mean is applied with the source-side raw valid-token mask. Decoder
counterfactuals, atom localization, edge-score probes, and dosage interventions
do not use this pooled vector: they keep the token state, apply the
decomposition under the active-target mask, and pass the resulting token tensor through
the frozen decoder.

\begin{table}[H]
\centering
\scriptsize
\setlength{\tabcolsep}{4pt}
\caption{Audit state definitions.}
\label{tab:audit_state_defs_app}
\begin{tabular}{p{0.16\linewidth}p{0.44\linewidth}p{0.20\linewidth}}
\toprule
State & Definition & Uses GT endpoint? \\
\midrule
$S_0=H_0$ & Raw encoded source endpoint, $\mathrm{Enc}(G_{\mathrm{source}})$ & No \\
$\widetilde H_1=F_\theta(S_0;e_d)$ & Shared attention-style updated state before task adaptation & No \\
$S_1=A_d(\widetilde H_1;e_d)$ / full & Natural transformed source state, aligned to the encoded target & No \\
$z_{\mathrm{target}}$ & Encoded ground-truth endpoint latent & Yes \\
$g=z_{\mathrm{target}}-S_0$ & Endpoint direction used for projection & Yes \\
$\Delta z=S_1-S_0$ & Model update displacement & No \\
$\Delta z_{\parallel}$ & Projection of $\Delta z$ onto $g$ & Yes \\
$\Delta z_{\perp}$ & $\Delta z-\Delta z_{\parallel}$ & Yes \\
$S_{\mathrm{mid}}$ & $S_0+\Delta z_{\parallel}$, program-stage counterfactual & Yes \\
Residual-only & $S_0+\Delta z_{\perp}$ & Yes \\
$S(\alpha)$ & $S_{\mathrm{mid}}+\alpha\Delta z_{\perp}$ & Yes \\
\bottomrule
\end{tabular}
\end{table}

\subsection{Residual-Dosage Selector Details}
\label{app:adaptive_alpha_selector}

\paragraph{Residual-dosage sanity check.}
In addition to the main decoder-facing bond probes, we test whether the
residual direction behaves like a calibrated correction when only its scalar
magnitude is changed. Starting from
$S_{\mathrm{mid}}=S_0+\Delta z_{\parallel}$, we decode
$S(\alpha)=S_{\mathrm{mid}}+\alpha\Delta z_{\perp}$ for a small alpha grid.
Performance rises sharply from the endpoint-parallel state, peaks near the
learned residual scale, and mildly declines when over-dosed. We treat this as
supporting evidence for calibrated residual correction, not as the primary
mechanistic claim.

\begin{table}[H]
\centering
\scriptsize
\setlength{\tabcolsep}{4pt}
\caption{Retro dosage response for
$S(\alpha)=S_{\mathrm{mid}}+\alpha\Delta z_{\perp}$. The residual behaves like
a calibrated correction rather than an unbounded force. Top-1 accuracy
bootstrap CIs are reported in Appendix~\ref{app:bootstrap_ci}.}
\label{tab:retro_dosage}
\begin{tabular}{lrrrr}
\toprule
$\alpha$ & Top-1 & Edit F1 & Formed F1 & Center F1 \\
\midrule
0 & 0.0997 & 0.4998 & 0.1927 & 0.6046 \\
0.25 & 0.4590 & 0.8055 & 0.7448 & 0.8819 \\
0.5 & \textbf{0.6369} & \textbf{0.8881} & \textbf{0.8868} & \textbf{0.9420} \\
1 & \textbf{0.6441} & \textbf{0.8868} & \textbf{0.8868} & \textbf{0.9418} \\
2 & 0.6345 & 0.8852 & 0.8849 & 0.9410 \\
\bottomrule
\end{tabular}
\end{table}

The adaptive-alpha analysis is evaluated only after the audit-time
decomposition has constructed $S_{\mathrm{mid}}$ and the residual direction
$\Delta z_{\perp}$. For each test sample, we decode the fine grid
$S(\alpha)=S_{\mathrm{mid}}+\alpha\Delta z_{\perp}$ with
$\alpha\in\{0,0.125,\ldots,2\}$. The oracle selector reports the best grid
point under label-available exact-match evaluation and is therefore only an
upper bound.

The GT-free proxy rule in Table~\ref{tab:adaptive_alpha_rescue} uses no
ground-truth exact match, edit F1, center F1, GT edge overlap, oracle alpha, or
true failure label. It is GT-free only after the audit decomposition has already
constructed $S_{\mathrm{mid}}$ and $\Delta z_{\perp}$. The rule keeps
$\alpha=1$ unless the decoded alpha sweep passes proxy gates for a local dosage
change; among passing candidates it chooses the alpha with the largest proxy
improvement over $\alpha=1$.

The thresholds are selected by a small grid sweep on a fixed 512-sample
sensitivity subset. For each candidate policy, we evaluate two label-available
quantities after selection: \emph{rescued}, the number of samples wrong at
$\alpha=1$ but correct after selection, and \emph{broken}, the number of
samples correct at $\alpha=1$ but wrong after selection. Candidate policies are
restricted to a conservative frontier with at most two broken samples on the
sensitivity subset and nonzero changed-sample fraction. We rank the remaining
rules by
\[
\mathrm{weighted\ net}_{b2}=\mathrm{rescued}-2\,\mathrm{broken},
\]
then by official accuracy and rescued count. Thus the selected thresholds are
tuned for cautious rescue, not for maximizing edit F1, center F1, or oracle
agreement. The final reported rule is one of the sensitivity-selected
GT-free proxy rules evaluated on the full retro test split.

\begin{table}[H]
\centering
\scriptsize
\setlength{\tabcolsep}{3pt}
\caption{Definitions used by the fine-grid dosage-rescue audit.}
\label{tab:adaptive_alpha_selector_defs_app}
\begin{tabular}{p{0.22\linewidth}p{0.68\linewidth}}
\toprule
Quantity & Definition \\
\midrule
Oracle alpha & For each sample, choose the alpha that maximizes official exact-match correctness; ties are broken by edit F1, center F1, distance to $\alpha=1$, and then alpha value. This is label-available and used only as an upper bound. \\
Oracle-rescuable & A sample that is wrong at $\alpha=1$ but correct for at least one alpha on the grid under the oracle rule. There are 247 such retro test samples. \\
Rescued / broken & Rescued means wrong at $\alpha=1$ and correct after the selector; broken means correct at $\alpha=1$ and wrong after the selector. \\
Reconstruction score & Negative mean rounding error between continuous bond logits and their rounded decoded bond orders over active atom pairs. Higher means the decoded graph is closer to a stable decoder boundary. \\
Predicted-edge margin & The weakest predicted changed-edge strength minus the strongest non-changed competitor strength. Higher means the decoded edit set is more separated from competing edits. \\
Consensus count & Number of alpha values that decode to the same edit signature. Higher means the candidate output is stable across nearby dosage values. \\
Internal alpha score & $r(\alpha)+0.05m(\alpha)+0.01\log(1+c(\alpha))-2I(\alpha)-0.5E_{\mathrm{src}}(\alpha)-0.2E_{\mathrm{zero}}(\alpha)-0.04|\alpha-1|$, where $r$ is reconstruction score, $m$ is predicted-edge margin, $c$ is consensus count, $I$ is invalid-output penalty, $E_{\mathrm{src}}$ is source-equal penalty, and $E_{\mathrm{zero}}$ is zero-edit penalty. The selector uses $\Delta$score relative to $\alpha=1$. \\
Proxy gates & A candidate alpha must have $\Delta$score $\ge -0.001$, be no less valid than $\alpha=1$, have consensus count at least 1, and have predicted-edge margin at least $-0.2$. Downward dosage is allowed only when the overdrive-risk score is in the top 60\%; upward dosage is allowed only when the under-refinement-risk score is in the top 60\% and edit-count explosion is at most 2. \\
Risk scores & Overdrive risk is the sum of residual-norm $z$-score, residual/perpendicular-ratio $z$-score, and $0.5$ times combined-instability $z$-score. Under-refinement risk is the residual/perpendicular-ratio $z$-score plus $0.5$ times residual-norm $z$-score plus $0.25$ times residual-top1-mass $z$-score. \\
Reported selector & The reported rule has score threshold $-0.001$, minimum consensus 1, risk quantile 0.4, maximum program-atypicality $z$-score 10, maximum upward edit-count explosion 2, and margin threshold $-0.2$. On the full retro test split, it changes 3.42\% of samples, rescues 46, breaks 2, and reaches official accuracy 0.6529. \\
\bottomrule
\end{tabular}
\end{table}

\begin{table}[H]
\centering
\scriptsize
\setlength{\tabcolsep}{3pt}
\caption{Latent aggregation domains used by Section~4 audits.}
\label{tab:audit_pooling_domains_app}
\begin{tabular}{p{0.28\linewidth}p{0.25\linewidth}p{0.34\linewidth}}
\toprule
Audit family & Latent domain & Notes \\
\midrule
Endpoint role readout & Graph-level pooled & Cached pooled $S_0$ or $z_{\mathrm{target}}$ with linear ridge probes \\
Update-object alignment and endpoint geometry & Graph-level pooled & Uses $\mathrm{Pool}_m(S_0)$, $\mathrm{Pool}_m(S_1)$, and $\mathrm{Pool}_m(z_{\mathrm{target}})$ \\
Stage/component class readout & Graph-level pooled & Encoder stages use raw valid-token pooling; decomposition components use active-target-mask pooling \\
Prototype, nearest-neighbor, subspace, and program-centroid analyses & Graph-level pooled & Class means/centroids are fit in pooled $\Delta z_{\perp}$ or $\Delta z_{\parallel}$ space \\
Functional state decoding, residual-only, and dosage & Token-level tensor & Constructs $S_{\mathrm{mid}}$, residual-only, or $S(\alpha)$ in $B\times L\times D$ before decoding \\
Boundary crossing, TER, and $d\,\mathrm{Bond}/dz$ & Token/edge-level & Decoder scores are edge-level; directions are token tensors under the active-target mask \\
Norm controls and token/sample shuffles & Token-level tensor & Random or shuffled residual tensors are decoded; summaries report sample means \\
Atom center mass, gradients, suppress/rescue & Atom/token-level & Uses per-token norms/gradients over active atoms, not graph-pooled vectors \\
Failure taxonomy and adaptive dosage & Mixed derived proxies & Reuses token-level decoded states plus atom/edge/sample proxy metrics \\
\bottomrule
\end{tabular}
\end{table}

Table~\ref{tab:audit_metric_defs_app} defines the main metrics used in
Section~4. Unless otherwise stated, confidence intervals use sample-level
bootstrap resampling; edge-level quantities are first aggregated to sample
means when sample identifiers are available.

\begin{table}[H]
\centering
\scriptsize
\setlength{\tabcolsep}{3pt}
\caption{Audit metric definitions and caveats.}
\label{tab:audit_metric_defs_app}
\begin{tabular}{p{0.16\linewidth}p{0.44\linewidth}p{0.20\linewidth}}
\toprule
Metric & Definition / aggregation & Caveat \\
\midrule
Official accuracy & Mean deterministic top-1 exact-match correctness & Not beam search \\
Edit F1 & Sample-level F1 over edited atom pairs / bond edits & Tensor proxy \\
Formed F1 & Sample-level F1 over formed bonds & Empty-set convention follows local implementation \\
Broken F1 & Sample-level F1 over broken bonds & Changed-order edits tracked separately \\
Center F1 & Sample-level F1 over atoms incident to bond edits & Reaction-center proxy \\
Center mass ratio & Residual token-norm mass on center atoms / active mass & Model-side localization \\
Boundary crossing & Fraction of selected edges crossing rounded/clamped decode thresholds & Decoder-score proxy \\
Competing-edge suppression & Suppression of high-score non-GT candidate edges & Candidate definition is automatic \\
Target-error reduction & Reduction in continuous $|\mathrm{score}-\mathrm{target}|$ & Not calibrated probability \\
$d\,\mathrm{Bond}/dz$ derivative & Local gradient directional derivative of bond score & Model sensitivity, not physical causality \\
Program atypicality & Distance of $\Delta z_{\parallel}$ to train-set family centroid & Correlational uncertainty proxy \\
Failure modes & Automatic endpoint/latent taxonomy labels on wrong samples & Not human chemistry labels \\
\bottomrule
\end{tabular}
\end{table}

\begin{table}[H]
\centering
\scriptsize
\setlength{\tabcolsep}{5pt}
\caption{Supplementary endpoint-geometry aggregate omitted from the main
table. The norm ratio is
$\mathrm{mean}(\|\Delta z_\perp\|)/\mathrm{mean}(\|\Delta z\|)$, a
norm-weighted counterpart to the per-sample mean residual fraction reported in
Table~\ref{tab:endpoint_geometry}.}
\label{tab:endpoint_geometry_norm_ratio_app}
\begin{tabular}{lr}
\toprule
Direction & Norm ratio \\
\midrule
Forward & 0.9307 \\
Retro & 0.6859 \\
\bottomrule
\end{tabular}
\end{table}
\subsection{Effective-Dimension Check for Directional Asymmetry}
\label{app:effective_dimension}

{As supporting geometry for the directional-asymmetry discussion in
Section~4, we compare the transformed endpoint state $S_1$ in forward and
retro on the full USPTO-50K test split. Global metrics use the covariance
spectrum of centered graph-level features; mean local intrinsic dimension uses
the Levina--Bickel kNN estimator with $k=20$.}

\begin{table}[H]
\centering
\scriptsize
\setlength{\tabcolsep}{5pt}
\caption{{Effective-dimension evidence for the transformed endpoint
state. Forward $S_1$ is more compact than retro $S_1$ by effective rank,
participation ratio, and mean local intrinsic dimension.}}
\label{tab:effective_dimension_direction_app}
\begin{tabular}{lrrrr}
\toprule
State & Effective rank & Participation ratio & Top-1/trace & Mean local ID \\
\midrule
Forward $S_1$ & 8.21 & 3.07 & 0.5585 & 12.60 \\
Retro $S_1$ & 11.25 & 5.66 & 0.2996 & 13.22 \\
\bottomrule
\end{tabular}
\end{table}

{These dimension metrics are used only as supporting evidence for task
asymmetry: lower effective dimension does not by itself imply easier chemistry
or weaker semantic content.}

\subsection{Artifact-Control Diagnostic Proxies}
\label{app:artifact_diagnostics}

\begin{table}[H]
\centering
\scriptsize
\setlength{\tabcolsep}{4pt}
\caption{Available diagnostic proxy details for the residual artifact controls in
Table~\ref{tab:artifact_controls}. ``Cross'' denotes decoder-boundary crossing
as defined in Section~4.}
\label{tab:artifact_diagnostics_app}
\begin{tabular}{llrr}
\toprule
Control & Variant & Edit F1 & Cross \\
\midrule
Reference & Full $S_1$ & 0.8868 & 0.5217 \\
\midrule
Norm-only random & Global norm & 0.5020 & 0.0163 \\
Norm-only random & Per-token norm & 0.5071 & 0.0288 \\
\midrule
Prototype & Class subspace & \multicolumn{2}{c}{aggregate-only} \\
Prototype & Class nearest neighbor & \multicolumn{2}{c}{aggregate-only} \\
Prototype & Class mean & \multicolumn{2}{c}{aggregate-only} \\
\bottomrule
\end{tabular}
\end{table}

\subsection{Detailed Section~4 Audit Protocols}
\label{app:section4_audit_details}

Tables~\ref{tab:substrate_geometry_protocol_app}--\ref{tab:failure_protocol_app}
spell out the aggregation domain used by each Section~4 audit.  The important distinction is that graph-level readouts are
masked means over latent tokens, while decoded counterfactuals, bond-boundary
probes, dosage sweeps, and atom-localization audits keep the full token tensor
until after decoder evaluation.

\begin{table}[H]
\centering
\scriptsize
\setlength{\tabcolsep}{3pt}
\caption{Detailed protocols for substrate, geometry, and staged decoding audits.}
\label{tab:substrate_geometry_protocol_app}
\begin{tabular}{p{0.18\linewidth}p{0.21\linewidth}p{0.20\linewidth}p{0.29\linewidth}}
\toprule
Audit & Implementation family & Latent domain & Protocol / reported quantity \\
\midrule
Endpoint performance context & Main endpoint evaluation & Decoded graph output & Reports deterministic top-1 endpoint exact match for the trained joint checkpoint; no latent pooling enters the score. \\
Opposite-endpoint role readout & role probe & Graph-level pooled & Fits linear/ridge role probes on cached pooled $S_0$ or $z_{\mathrm{target}}$ vectors to test whether endpoint role information is readable. \\
Update-object alignment & delta-z analysis & Graph-level pooled & Computes $\mathrm{Pool}_m(S_0)$, $\mathrm{Pool}_m(S_1)$, $\mathrm{Pool}_m(z_{\mathrm{target}})$, then reports cosine/projection/orthogonal residual geometry. \\
Endpoint geometry & delta-z analysis & Graph-level pooled & Uses the same pooled endpoint vectors to measure distance-to-target, parallel fraction, residual norm share, and related aggregate geometry. \\
Staged decoding & selective ablation & Token-level tensor & Builds $S_0$, $S_{\mathrm{mid}}$, residual-only, and $S_1$ as $B\times L\times D$ tensors under the active-target mask, decodes each with the frozen residual decoder, and reports official/edit/formed/broken/center metrics. \\
Target-encoder oracle & oracle decoding & Token-level tensor & Replaces the transformed endpoint state by $z_{\mathrm{target}}=\mathrm{Enc}(G_{\mathrm{target}})$ before the same frozen decoder to test whether target encoder latents are directly decoder-executable. \\
Functional split & selective ablation & Token-level tensor & Compares full, endpoint-parallel, and endpoint-orthogonal counterfactual states after decoding; no pooled vector is fed to the decoder. \\
Parallel preconditioner & parallel dosage & Token-level tensor & Keeps the true residual fixed and varies the endpoint-parallel component before decoding to test whether the residual needs a coarse endpoint-conditioned state. \\
\bottomrule
\end{tabular}
\end{table}

\begin{table}[H]
\centering
\scriptsize
\setlength{\tabcolsep}{3pt}
\caption{Detailed protocols for residual controls and decoder-grounding audits.}
\label{tab:residual_decoder_protocol_app}
\begin{tabular}{p{0.18\linewidth}p{0.21\linewidth}p{0.20\linewidth}p{0.29\linewidth}}
\toprule
Audit & Implementation family & Latent domain & Protocol / reported quantity \\
\midrule
Norm-matched random controls & selective ablation & Token-level tensor & Replaces $\Delta z_{\perp}$ by random directions scaled to the same masked norm, decodes the resulting token state, and reports endpoint/edit proxies. \\
Token/sample/class shuffles & shuffle controls & Token-level tensor for decoding & Permutes or substitutes residual token tensors before decoding; graph-level class prototypes may be fit for selection, but the decoded intervention itself is token-level. \\
Class prototype controls & prototype controls & Mixed & Fits class mean/nearest-neighbor/subspace controls in pooled residual space, then evaluates decoded prototype-derived token controls where available. \\
Boundary crossing & bond-boundary audit & Token/edge-level & Evaluates continuous decoder bond scores on selected edges and counts threshold crossings caused by endpoint-parallel versus residual movement. \\
Score delta and TER & bond-boundary audit & Token/edge-level & Computes score changes and target-error reduction on decoder edge scores after decoding counterfactual token states. \\
Directional derivative & decoder-gradient audit & Token/edge-level & Uses masked token-space directions and decoder gradients to measure local $d\,\mathrm{Bond}/dz$ alignment for target and competing edges. \\
Residual dosage & dosage response & Token-level tensor & Sweeps $S(\alpha)=S_{\mathrm{mid}}+\alpha\Delta z_{\perp}$ before decoding and summarizes official/edit/center response curves. \\
\bottomrule
\end{tabular}
\end{table}

\begin{table}[H]
\centering
\scriptsize
\setlength{\tabcolsep}{3pt}
\caption{Detailed protocols for semantic organization, atom localization, and failure diagnosis.}
\label{tab:failure_protocol_app}
\begin{tabular}{p{0.18\linewidth}p{0.21\linewidth}p{0.20\linewidth}p{0.29\linewidth}}
\toprule
Audit & Implementation family & Latent domain & Protocol / reported quantity \\
\midrule
Encoder-stage class readout & layerwise readout & Graph-level pooled & Pools each encoder layer over raw valid source tokens and fits cross-validated linear reaction-class probes. \\
Component class readout & component readout & Graph-level pooled & Pools $S_0$, $S_1$, $\Delta z$, $\Delta z_{\parallel}$, and $\Delta z_{\perp}$ under the active-target mask before linear probing. \\
Program centroid / atypicality & program centroid & Graph-level pooled & Fits class/family centroids in pooled $\Delta z_{\parallel}$ or residual space and relates distance to residual demand or failure. \\
Atom center mass & node residual audit & Atom/token-level & Computes per-token residual norm mass on active atoms and compares mass on reaction-center atoms against total active mass. \\
Atom gradients and suppress/rescue & atom/gradient audit & Atom/token-level & Uses per-token gradients, residual norms, and atom suppress/rescue interventions; no graph-level latent vector is used for the atom proxy itself. \\
Merged failure taxonomy & failure taxonomy & Mixed derived proxies & Combines decoded state metrics, dosage response, atom center mass, edge competition, and program atypicality into automatic endpoint/latent failure labels. \\
Weak alpha rescue & adaptive dosage & Token-level plus derived selector & Applies residual dosage choices to token-level decoded alpha sweeps; GT-free selectors are GT-free only after the audit decomposition has been constructed. \\
\bottomrule
\end{tabular}
\end{table}

\subsection{Bootstrap Confidence Intervals for Key Audit Results}
\label{app:bootstrap_ci}

Tables~\ref{tab:bootstrap_stage_app}--\ref{tab:bootstrap_failure_alpha_app}
report bootstrap intervals for the main sample-level claims in Section~4. We
resample test samples with replacement; edge-level quantities are first
aggregated to sample means whenever sample identifiers are available. Entries
marked aggregate-only in the main text are not assigned invented intervals.

\begin{table}[H]
\centering
\scriptsize
\setlength{\tabcolsep}{4pt}
\caption{Bootstrap CIs for staged decoding and functional split.}
\label{tab:bootstrap_stage_app}
\begin{tabular}{llr}
\toprule
Quantity & Metric & Mean [95\% CI] \\
\midrule
Retro $S_0$ & Official & \ci{0.0405}{0.0350}{0.0463} \\
Retro $S_{\mathrm{mid}}$ & Official & \ci{0.0997}{0.0913}{0.1075} \\
Retro $S_1$ & Official & \ci{0.6441}{0.6305}{0.6571} \\
Retro residual-only & Official & \ci{0.6303}{0.6171}{0.6437} \\
\midrule
Forward $S_0$ & Official & \ci{0.8694}{0.8602}{0.8784} \\
Forward $S_{\mathrm{mid}}$ & Official & \ci{0.8816}{0.8724}{0.8902} \\
Forward $S_1$ & Official & \ci{0.8832}{0.8744}{0.8926} \\
\bottomrule
\end{tabular}
\end{table}

\begin{table}[H]
\centering
\scriptsize
\setlength{\tabcolsep}{4pt}
\caption{Bootstrap CIs for residual controls, decoder grounding, and atom-level
organization.}
\label{tab:bootstrap_mechanism_app}
\begin{tabular}{llr}
\toprule
Family & Quantity & Mean [95\% CI] \\
\midrule
Residual control & True residual official & \ci{0.6441}{0.6301}{0.6563} \\
Residual control & Random global-norm r0 official & \ci{0.1029}{0.0949}{0.1112} \\
Residual control & Random per-token-norm r0 official & \ci{0.1051}{0.0969}{0.1138} \\
Residual control & Token-shuffle r0 official & \ci{0.0839}{0.0765}{0.0917} \\
Residual control & Sample-shuffle r0 official & \ci{0.1176}{0.1084}{0.1270} \\
Residual control & Same-class shuffle r0 official & \ci{0.1190}{0.1104}{0.1280} \\
\midrule
Decoder boundary & {Aggregate edit residual crossing} & {\ci{0.5217}{0.5081}{0.5353}} \\
Decoder boundary & {Aggregate formed residual crossing} & {\ci{0.6054}{0.5898}{0.6208}} \\
Decoder boundary & {Aggregate wrong-edge crossing} & {\ci{0.0004}{0.0004}{0.0005}} \\
Decoder boundary & {Aggregate competing-edge suppression} & {\ci{0.8525}{0.8302}{0.8732}} \\
\midrule
$d\,\mathrm{Bond}/dz$ & {Aggregate $\Delta z_{\perp}$ on GT formed edges} & {\ci{+1.0401}{+0.9134}{+1.1767}} \\
$d\,\mathrm{Bond}/dz$ & {Aggregate $\Delta z_{\parallel}$ on GT formed edges} & {\ci{+0.1839}{+0.1566}{+0.2137}} \\
$d\,\mathrm{Bond}/dz$ & {Aggregate random on GT formed edges} & {\ci{-0.0017}{-0.0085}{+0.0050}} \\
$d\,\mathrm{Bond}/dz$ & {Aggregate $\Delta z_{\perp}$ suppressing competing edges} & {\ci{-0.4104}{-0.5126}{-0.3219}} \\
\midrule
Atom localization & Retro $\Delta H_{\perp}$ center mass & \ci{0.3255}{0.3210}{0.3300} \\
Atom localization & Retro $\Delta H_{\parallel}$ center mass & \ci{0.4918}{0.4875}{0.4961} \\
Atom function & Top-3 suppress formed F1 drop & \ci{0.6112}{0.5972}{0.6245} \\
Atom function & Top-3 rescue formed F1 gain & \ci{0.2960}{0.2834}{0.3081} \\
\bottomrule
\end{tabular}
\end{table}

\begin{table}[H]
\centering
\scriptsize
\setlength{\tabcolsep}{4pt}
\caption{Bootstrap CIs for residual dosage, failure taxonomy, and weak alpha
selection.}
\label{tab:bootstrap_failure_alpha_app}
\begin{tabular}{llr}
\toprule
Family & Quantity & Mean [95\% CI] \\
\midrule
Dosage & $\alpha=0$ official & \ci{0.0997}{0.0917}{0.1079} \\
Dosage & $\alpha=0.25$ official & \ci{0.4590}{0.4452}{0.4737} \\
Dosage & $\alpha=0.5$ official & \ci{0.6369}{0.6231}{0.6495} \\
Dosage & $\alpha=1$ official & \ci{0.6441}{0.6303}{0.6561} \\
Dosage & $\alpha=2$ official & \ci{0.6345}{0.6209}{0.6481} \\
\midrule
Failure diagonal & Program $\rightarrow$ Program/Parallel & \ci{0.858}{0.810}{0.903} \\
Failure diagonal & Dosage $\rightarrow$ Dosage & \ci{0.886}{0.771}{0.971} \\
Failure diagonal & Decoder $\rightarrow$ Decoder realization & \ci{0.969}{0.906}{1.000} \\
Failure off-diagonal & Refinement $\rightarrow$ Program/Parallel & \ci{0.413}{0.387}{0.440} \\
Failure off-diagonal & Nonresponsive $\rightarrow$ Residual failure & \ci{0.504}{0.445}{0.555} \\
\midrule
Alpha selection & Fixed $\alpha=1$ official & \ci{0.6441}{0.6303}{0.6581} \\
Alpha selection & Oracle alpha official & \ci{0.6934}{0.6810}{0.7062} \\
Alpha selection & Conservative selector official & \ci{0.6529}{0.6395}{0.6659} \\
\bottomrule
\end{tabular}
\end{table}

\subsection{Failure-Diagnosis Visualization}
\label{app:failure_diagnosis_viz}
\vspace{-0.5in}
\begin{figure}[H]
\centering
\includegraphics[width=\linewidth]{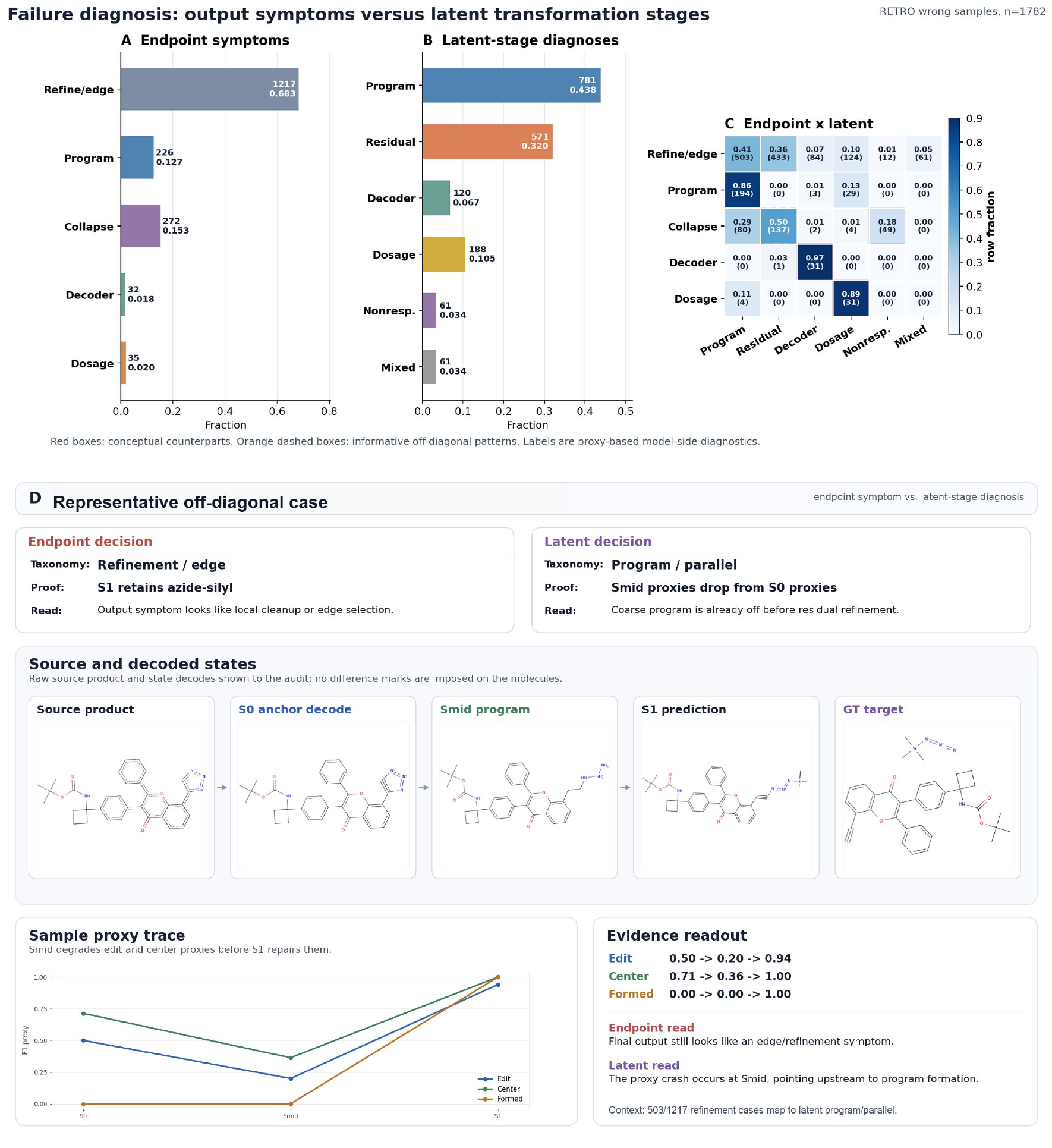}
\caption{Two-level failure diagnosis on wrong retro samples ($n=1782$).
Endpoint symptoms and latent-stage diagnoses agree on the diagonal, while
off-diagonal cases reveal hidden program/parallel failures behind local-looking
refinement errors.}
\label{fig:merged_failure_taxonomy}
\end{figure}

\subsection{Qualitative Visualization of Update-Component Semantics}
\label{app:delta_space_viz}

\begin{figure}[H]
\centering
\includegraphics[width=\linewidth]{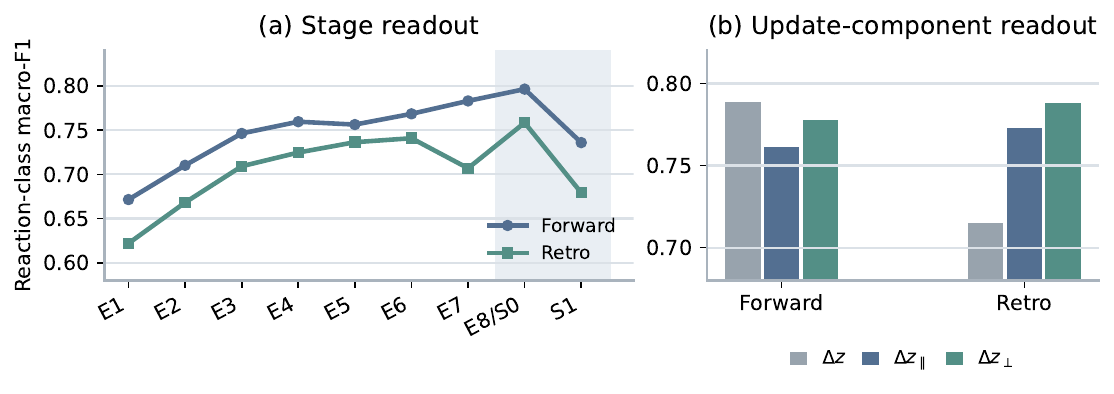}
\caption{Reaction-class linear readouts across stages and update components.
(a) Macro-F1 mostly builds through encoder layers up to E8/$S_0$ and drops after the
UniRxnAxis update to $S_1$. (b) Using the same 5-fold ridge-probe protocol on the
active-target-mask pooled decomposition states, retro component separation
increases reaction-class readability relative to the raw update, whereas forward
$\Delta z$ remains slightly higher than either component.}
\label{fig:stage_class_readout}
\end{figure}

\begin{figure}[H]
\centering
\includegraphics[width=\linewidth]{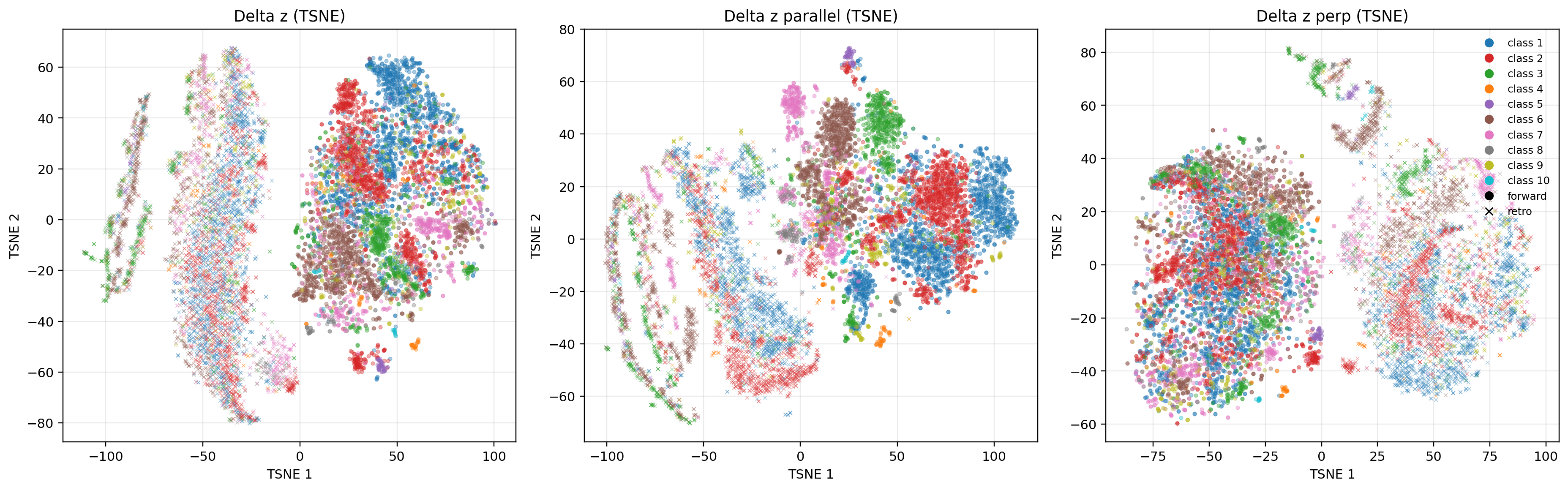}
\caption{{Qualitative joint t-SNE visualization of forward and retro update
components colored by reaction class. Coordinates are fit jointly over both
directions within each component panel; circles denote forward examples and
crosses denote retro examples. The visualization is appendix-only and is not
used as primary evidence.}}
\label{fig:joint_forward_retro_delta_tsne_app}
\end{figure}

{Figure~\ref{fig:joint_forward_retro_delta_tsne_app} visualizes forward and retro
update components in a jointly fit t-SNE space and colors points by reaction
class. The plot is consistent with the linear readout evidence in
Figure~\ref{fig:stage_class_readout}: in full-dimensional probes,
$\Delta z_{\perp}$ is more readable than $\Delta z_{\parallel}$ for both
reaction class (macro-F1 $0.762$ vs.\ $0.288$) and bond-edit bucket (macro-F1
$0.687$ vs.\ $0.451$). We use this figure only as a qualitative semantic
signal, since t-SNE layouts depend on visualization choices and do not
establish mechanism by themselves.}

\subsection{Atom-Level and Sample-Level Supporting Diagnostics}
\label{app:atom_sample_diagnostics}

Chemistry alignment should also appear in atom-level coordinates, especially
around reaction centers. Under the raw-anchor convention, retro
$\Delta H_{\parallel}$ carries the largest reaction-center mass, while
$\Delta H_{\perp}$ remains strongly center-enriched and almost always includes
a reaction-center atom among its top residual atoms
(Table~\ref{tab:atom_level_organization}, top). Thus the atom-level result is
not ``residual mass is uniquely most localized''; it is that both endpoint
alignment and residual refinement concentrate around chemically edited atoms,
with the residual retaining localized structure after removing the
endpoint-reference component.

Localization alone could be a visual prior, so we test whether the localized
atoms matter to the decoder. Decoder gradients concentrate on reaction-center
atoms and overlap with high-residual atoms. Suppressing top residual atoms
causes large edit, formed-bond, and center F1 drops; rescuing those atoms
restores part of the performance (Table~\ref{tab:atom_level_organization}, bottom). Thus residual
localization is functionally connected to bond-edit computation.

\begin{table}[H]
\centering
\scriptsize
\setlength{\tabcolsep}{4pt}
\caption{Atom-level organization of emergent steering. Top: both
endpoint-parallel and residual components concentrate mass on reaction-center
atoms, with the endpoint-parallel component carrying the largest center mass
under the raw-anchor convention. Bottom: localized residual atoms are
functionally connected to decoder bond-edit computation. Bootstrap CIs are
reported in Appendix~\ref{app:bootstrap_ci}.}
\label{tab:atom_level_organization}
\label{tab:atom_center_mass}
\label{tab:atom_function}
\begin{tabular}{llr}
\toprule
\multicolumn{3}{l}{\textit{Reaction-center localization}} \\
\midrule
Direction & Metric & Value \\
\midrule
Retro & $\Delta H_{\parallel}$ center mass & \textbf{0.4918} \\
Retro & $\Delta H_{\perp}$ center mass & 0.3255 \\
Retro & $\Delta H_{\mathrm{total}}$ center mass & 0.4075 \\
Retro & Label fraction & 0.1596 \\
Forward & $\Delta H_{\parallel}$ center mass & \textbf{0.2340} \\
Forward & $\Delta H_{\perp}$ center mass & 0.1526 \\
Forward & $\Delta H_{\mathrm{total}}$ center mass & 0.1928 \\
\midrule
\multicolumn{3}{l}{\textit{Functional atom-level evidence}} \\
\midrule
Probe & Metric & Value \\
\midrule
Gradient & Center mass & \textbf{0.5897} \\
Gradient & Spearman grad norm vs residual norm & \textbf{0.4549} \\
Gradient & Top-3 gradient/residual atom overlap & \textbf{0.6491} \\
\midrule
Suppress & Edit F1 drop & 0.3340 \\
Suppress & Formed F1 drop & \textbf{0.6112} \\
Suppress & Center F1 drop & 0.2641 \\
Rescue & Edit F1 gain & 0.1436 \\
Rescue & Formed F1 gain & \textbf{0.2960} \\
Rescue & Center F1 gain & 0.1581 \\
\bottomrule
\end{tabular}
\end{table}

The parallel/residual division is also visible at the sample level. If
$\Delta z_{\parallel}$ provides a coarse program, atypical coarse programs
should require more residual refinement. We define program atypicality as the
distance from a sample's $\Delta z_{\parallel}$ to the training centroid for
its reaction class. Larger distance predicts a larger residual-to-parallel norm
ratio and slightly worse $S_{\mathrm{mid}}$ edit and center quality, but it is
not a strong predictor of final correctness or residual-stage gain
(Table~\ref{tab:sample_atypicality}). This weakens the old ``program
atypicality explains residual demand'' story: the coarse program geometry is a
useful sample-level covariate, not a sufficient account of residual content.

\begin{table}[H]
\centering
\small
\setlength{\tabcolsep}{5pt}
\caption{Program atypicality is a weak residual-demand covariate. Values are
Spearman correlations using reaction-class centroids of
$\Delta z_{\parallel}$.}
\label{tab:sample_atypicality}
\begin{tabular}{lr}
\toprule
Correlation & Spearman \\
\midrule
Distance vs $\|\Delta z_{\perp}\|$ & 0.2161 \\
Distance vs $\|\Delta z_{\perp}\|/\|\Delta z_{\parallel}\|$ & \textbf{0.3754} \\
Distance vs $S_{\mathrm{mid}}$ edit F1 & -0.0387 \\
Distance vs $S_{\mathrm{mid}}$ center F1 & -0.0492 \\
Distance vs $S_1-S_{\mathrm{mid}}$ edit gain & -0.0738 \\
Distance vs $S_1-S_{\mathrm{mid}}$ center gain & -0.0399 \\
\bottomrule
\end{tabular}
\end{table}

\subsection{Audited UniRxnAxis Checkpoint Configuration}

Table~\ref{tab:urf_checkpoint_config_app} lists the implementation details of
the checkpoint used for all directional audits in Section~4. These details are
not needed to understand the method, but they define the audited model instance.

\begin{table}[h]
\centering
\small
\setlength{\tabcolsep}{5pt}
\caption{UniRxnAxis checkpoint used in the directional audit. The model is trained
jointly on forward and retrosynthetic directions; reaction-class conditioning
and cycle loss are disabled in this checkpoint.}
\label{tab:urf_checkpoint_config_app}
\begin{tabular}{p{0.30\linewidth}p{0.58\linewidth}}
\toprule
Component & Configuration \\
\midrule
Latent/token dimensions & $D=256$, $L=128$ atom slots \\
Backbone encoder/decoder & 8 Transformer layers, 8 heads, dropout 0.1 \\
Shared update $F_\theta$ & $K=1$, shared parameters, 2 layers, hidden dim 512, 8 heads \\
Direction conditioning & learned embedding $e_d$, retro=0 and forward=1 \\
Task adapter $A_d$ & direction-conditioned 2-layer attention adapter applied to $\widetilde H_1$ in the supervised transformation \\
Decoder input & fuse mode, $f_{\mathrm{fuse}}([S_0\mathbin{\|}S_1-S_0];e_d)$ before residual graph decoding \\
Bond decoder & direction-conditioned residual bond-change decoder, depth 3, FF multiplier 2 \\
Endpoint alignment weight & $\lambda_{\mathrm{align}}=0.1$ \\
Training mixture & forward/retro ratio $0.5/0.5$ \\
Training hardware & $4\times$ NVIDIA A10, ${\approx}10$ h (${\approx}40$ GPU-hours) \\
\bottomrule
\end{tabular}
\end{table}

\subsection{Endpoint-Parallel Preconditioning Controls}

Table~\ref{tab:parallel_dosage_app} reports the full dosage sweep used to test
the conditional contribution of $\Delta z_{\parallel}$ when the true residual
steering direction is fixed. The state is
$S(\beta)=S_0+\beta\Delta z_{\parallel}+\Delta z_{\perp}$; thus $\beta=0$ is
the residual-only intervention, $\beta=1$ is the full learned state, and
negative $\beta$ reverses the endpoint-parallel component.

\begin{table}[H]
\centering
\small
\setlength{\tabcolsep}{5pt}
\caption{Endpoint-parallel dosage under fixed true residual steering.}
\label{tab:parallel_dosage_app}
\begin{tabular}{lrrrr}
\toprule
$\beta$ & Top-1 & Edit F1 & Formed F1 & Center F1 \\
\midrule
-1.0 & 0.3843 & 0.4346 & 0.2219 & 0.6276 \\
-0.5 & 0.5482 & 0.7101 & 0.5816 & 0.8242 \\
0.0 & 0.6303 & 0.8799 & 0.8761 & 0.9371 \\
0.25 & 0.6415 & 0.8867 & 0.8879 & 0.9416 \\
0.5 & 0.6423 & 0.8870 & 0.8877 & 0.9413 \\
1.0 & \textbf{0.6441} & 0.8868 & 0.8868 & 0.9418 \\
1.5 & 0.6425 & \textbf{0.8885} & \textbf{0.8879} & \textbf{0.9429} \\
2.0 & 0.6431 & 0.8880 & 0.8872 & 0.9428 \\
\bottomrule
\end{tabular}
\end{table}

Table~\ref{tab:parallel_controls_app} compares the true endpoint-parallel
component against controls that preserve coarse statistics or empirical
vectors while breaking the sample-specific endpoint-aligned direction. The
true parallel component improves over residual-only, while random and shuffled
controls remain below it.

\begin{table}[H]
\centering
\small
\setlength{\tabcolsep}{5pt}
\caption{Endpoint-parallel controls under fixed true residual steering.}
\label{tab:parallel_controls_app}
\label{tab:parallel_preconditioner}
\begin{tabular}{lrrrr}
\toprule
Variant & Top-1 & Edit F1 & Formed F1 & Center F1 \\
\midrule
True parallel & \textbf{0.6441} & \textbf{0.8868} & \textbf{0.8868} & \textbf{0.9418} \\
Residual-only & 0.6303 & 0.8799 & 0.8761 & 0.9371 \\
Anti-parallel & 0.3843 & 0.4346 & 0.2219 & 0.6276 \\
Random global-norm parallel & 0.6275 & 0.8782 & 0.8733 & 0.9358 \\
Random per-token-norm parallel & 0.6209 & 0.8694 & 0.8598 & 0.9303 \\
Sample-shuffled parallel & 0.6089 & 0.8670 & 0.8593 & 0.9302 \\
Same-class shuffled parallel & 0.5990 & 0.8660 & 0.8579 & 0.9293 \\
Token-shuffled parallel & 0.5690 & 0.8528 & 0.8436 & 0.9254 \\
\bottomrule
\end{tabular}
\end{table}
\subsection{Failure Taxonomy Definitions}
\label{app:failure_taxonomy_definitions}

The failure taxonomy is an automatic proxy audit over retro test examples that
are wrong under the natural $\alpha=1$ decoded output. It is not a human
chemical-error adjudication. Endpoint labels describe the decoded output
symptom, while latent labels describe which part of the
$S_0\!\rightarrow\!S_{\mathrm{mid}}\!\rightarrow\!S_1$ trace appears
unreliable. Multiple fine proxy triggers may fire for one sample; the reported
merged label groups related triggers into the coarse categories below, with
mixed/ambiguous used when no single coarse explanation is dominant.

\begin{table}[H]
\centering
\scriptsize
\setlength{\tabcolsep}{3pt}
\caption{Coarse failure-taxonomy labels used by the endpoint--latent crosstab.}
\label{tab:failure_taxonomy_definitions_app}
\begin{tabular}{p{0.18\linewidth}p{0.24\linewidth}p{0.46\linewidth}}
\toprule
Axis & Coarse label & Meaning / typical proxy evidence \\
\midrule
\multirow{5}{=}{Endpoint symptom}
& Refinement / edge-selection & The final decoded graph has local-looking edit errors, such as a wrong competing edge, center mislocalization, or an incomplete final edit. \\
& Program & The decoded output appears wrong at a broad reaction-program level, beyond a single final edge choice. \\
& Nonresponsive / collapse & The decoded output barely responds to the residual path, collapses to a scaffold-like output, or remains near an unedited state. \\
& Decoder realization & The latent/bond proxies look plausible, but the discrete decoder still misses exact endpoint realization. \\
& Dosage & Changing the residual magnitude appears to improve the proxy path, suggesting under- or over-application of the residual. \\
\midrule
\multirow{6}{=}{Latent-stage diagnosis}
& Program / parallel & The program-stage trace is atypical, undercommitted, mislocalized, or poorly anchored before residual refinement. \\
& Residual refinement & $S_{\mathrm{mid}}$ is partly plausible, but the residual direction is insufficient, mislocalized, or points toward a competing edit. \\
& Decoder realization & Latent localization and edit proxies are plausible, but exact graph realization fails after decoding. \\
& Dosage & The residual direction appears useful but the learned scalar amount is too weak, too strong, or otherwise off its best local range. \\
& Nonresponsive / off-manifold & Sweeping residual dosage does not produce a useful decoded response, suggesting the local latent path is ineffective. \\
& Mixed / ambiguous & Multiple coarse latent explanations fire without a dominant assignment. \\
\bottomrule
\end{tabular}
\end{table}

\paragraph{Concrete thresholding convention.}
The fine labels below use fixed, quantile-based thresholds computed over the
available USPTO-50K retro test rows before filtering to wrong examples
($n=5007$ unless noted). We write $\mathrm{low}(x)$ for values in the bottom
quartile, $\mathrm{very\_low}(x)$ for the bottom decile,
$\mathrm{high}(x)$ for the top quartile, and
$\mathrm{very\_high}(x)$ for the top decile. When a low cutoff equals the
observed maximum, as happens for saturated F1 proxies, the rule uses strict
$x<q_{25}$ so that perfect F1 values are not labeled low only because the
distribution is saturated. Rules fire only on wrong natural $\alpha=1$
outputs; correct examples are tracked only as diagnostic warnings. If a proxy
is unavailable for a sample, the corresponding evidence item simply does not
fire.

\begin{table}[H]
\centering
\scriptsize
\setlength{\tabcolsep}{3pt}
\caption{Numeric cutoffs used by the automatic failure-taxonomy rules. Low and
high correspond to $q_{25}$ and $q_{75}$; very-low and very-high correspond to
$q_{10}$ and $q_{90}$.}
\label{tab:failure_taxonomy_thresholds_app}
\resizebox{\linewidth}{!}{%
\begin{tabular}{lrrrrr}
\toprule
Proxy & $n$ & $q_{10}$ & $q_{25}$ & $q_{75}$ & $q_{90}$ \\
\midrule
$S_{\mathrm{mid}}$ edit F1 & 5007 & 0.0000 & 0.1292 & 0.6667 & 1.0000 \\
$S_{\mathrm{mid}}$ center F1 & 5007 & 0.0000 & 0.3077 & 0.8000 & 1.0000 \\
$S_1$ edit F1 & 5007 & 0.4000 & 1.0000 & 1.0000 & 1.0000 \\
$S_1$ center F1 & 5007 & 0.8000 & 1.0000 & 1.0000 & 1.0000 \\
$S_1-S_{\mathrm{mid}}$ edit-F1 gain & 5007 & 0.0000 & 0.1818 & 0.6000 & 1.0000 \\
$S_1-S_{\mathrm{mid}}$ center-F1 gain & 5007 & 0.0000 & 0.2000 & 0.5000 & 1.0000 \\
Residual edit/center gain & 5007 & 0.0000 & 0.2667 & 0.5303 & 1.0000 \\
Program atypicality & 5007 & 0.3055 & 0.3586 & 0.5584 & 0.7023 \\
$\|\Delta z_{\parallel}\|$ & 5007 & 0.0113 & 0.0148 & 0.0302 & 0.0534 \\
$\|\Delta z_{\perp}\|$ & 5007 & 0.0228 & 0.0246 & 0.0321 & 0.0391 \\
$\|\Delta z\|$ residual norm & 5007 & 0.2571 & 0.2840 & 0.4059 & 0.5779 \\
Center mass ratio & 5007 & 0.1730 & 0.2038 & 0.4000 & 0.5896 \\
Off-center mass ratio & 5007 & 0.4104 & 0.6000 & 0.7962 & 0.8270 \\
Residual entropy & 5007 & 0.9342 & 0.9501 & 0.9741 & 0.9794 \\
Mislocalized residual score & 5007 & 0.3838 & 0.5699 & 0.7752 & 0.8081 \\
Combined instability score & 5007 & 1.7549 & 2.1491 & 2.5664 & 2.6317 \\
Residual boundary-crossing rate & 5002 & 0.0000 & 0.0000 & 0.5000 & 0.9333 \\
Wrong-edge crossing rate & 686 & 0.0000 & 1.0000 & 1.0000 & 1.0000 \\
Competing-edge suppression & 686 & -0.1608 & 0.2962 & 0.7595 & 0.9319 \\
Residual target-error reduction & 5002 & 0.0231 & 0.1177 & 0.5325 & 0.8698 \\
$S_1$ edit-score margin & 5002 & -1.0034 & -0.0128 & -0.0067 & -0.0061 \\
Top non-GT edit magnitude & 5007 & 0.0026 & 0.0027 & 0.0033 & 0.6106 \\
Best dosage $\alpha$ & 5007 & -0.2500 & 0.2500 & 0.5000 & 0.5000 \\
Same-class residual-control edit F1 & 5007 & 0.0000 & 0.2222 & 0.7778 & 0.8889 \\
True minus same-class edit F1 & 5007 & 0.0000 & 0.1111 & 0.5556 & 0.9394 \\
GT-edit residual derivative & 1023 & -0.1357 & 0.0009 & 0.5611 & 1.0839 \\
GT-edit residual gradient cosine & 1023 & -0.0820 & -0.0128 & 0.1151 & 0.2095 \\
\bottomrule
\end{tabular}%
}
\end{table}

\begin{table}[H]
\centering
\scriptsize
\setlength{\tabcolsep}{3pt}
\caption{Endpoint-side fine triggers before coarse merging. All rows are
conditioned on the natural $S_1$ decoded output being wrong.}
\label{tab:endpoint_failure_rulebook_app}
\begin{tabular}{p{0.25\linewidth}p{0.67\linewidth}}
\toprule
Fine endpoint trigger & Thresholded proxy condition \\
\midrule
Program-level failure & Low $S_{\mathrm{mid}}$ edit F1 or center F1, together with high program atypicality or high family-centroid distance. \\
Center mislocalization & Low center mass ratio, high off-center mass ratio, or a top-3/top-5 residual-center miss. \\
Competing-edge failure & At least two of: low competing-edge suppression, high wrong-edge crossing, high top non-GT edit magnitude, or low $S_1$ edit-score margin. \\
Under-refinement & $\alpha=1$ proxy quality is low, and either best $\alpha>1$, $\alpha=2$ improves over $\alpha=1$, or the dosage sweep is monotonic-improving. \\
Overshoot & Dosage response is overshoot, and either best $\alpha<1$ or $\alpha=1$ is worse than a lower dose. \\
Nonresponsive / collapse & Dosage response is nonresponsive, and the stage path remains low, the alpha path is flat, or the alpha-sweep proxy range is unavailable. \\
Refinement-level failure & $S_{\mathrm{mid}}$ edit and center F1 are not low, but $S_1$ edit/center F1 or the $S_{\mathrm{mid}}\!\rightarrow\!S_1$ residual gain is low. \\
Template/prototype insufficiency & The true sample residual control is high, the same-class/bucket control is low, and the true-minus-control gap is high. \\
Program-atypical hard case & Program atypicality is very high, residual norm is high, and $S_{\mathrm{mid}}$ edit or center F1 is low. \\
Residual-instability risk & At least two of: high residual entropy, high off-center mass, low edit-score margin, high mislocalized residual score, or high combined instability score. \\
Untyped proxy-insufficient & No endpoint rule fires with enough available evidence; used only as a low-confidence fallback before coarse merging. \\
\bottomrule
\end{tabular}
\end{table}

The endpoint coarse merge maps program-level, program-atypical, and
template/prototype triggers to \emph{Program}; center, competing-edge, and
refinement-level triggers to \emph{Refinement / edge-selection}; under- and
over-dosage triggers to \emph{Dosage}; and nonresponsive triggers to
\emph{Nonresponsive / collapse}. \emph{Decoder realization} is reserved for
exact-output misses where the latent and decoder-score proxies are otherwise
plausible.

\begin{table}[H]
\centering
\scriptsize
\setlength{\tabcolsep}{3pt}
\caption{Latent-stage fine triggers before coarse merging. All rows are
conditioned on the natural $S_1$ decoded output being wrong.}
\label{tab:latent_failure_rulebook_app}
\begin{tabular}{p{0.25\linewidth}p{0.67\linewidth}}
\toprule
Fine latent trigger & Thresholded proxy condition \\
\midrule
Anchor mislocalization & $S_0$ edit F1 or center F1 is very low. This rule is marked limited-support because $S_0$ readout evidence is sparse. \\
Program atypicality & Program atypicality or family-centroid distance is high, and either $S_{\mathrm{mid}}$ edit/center F1 is low or residual norm is high. \\
Program undercommitment & At least two $S_{\mathrm{mid}}-S_0$ gains are low across edit, formed-bond, and center proxies, with low $\|\Delta z_{\parallel}\|$ or low $S_{\mathrm{mid}}$ quality. \\
Wrong program commitment & $S_{\mathrm{mid}}$ edit/center F1 is low, a competing/wrong-edge signal fires, the program proxy or $\|\Delta z_{\parallel}\|$ is high, and residual gain remains low. \\
Residual mislocalization & At least two of low center mass, high off-center mass, top-k center miss, or high residual entropy, paired with low $S_1$ center quality or low center mass. \\
Residual misdirection & At least two of low/negative GT-edit derivative, high wrong-edge crossing, low competing-edge suppression, or low target-error reduction, with low or negative residual gain. \\
Residual insufficiency & $S_{\mathrm{mid}}$ edit/center F1 is not very low, residual gain is low, and dosage or crossing evidence suggests that $\alpha=1$ is too weak. \\
Residual overdrive & Dosage response is overshoot, and best $\alpha<1$, $\alpha=1$ is worse than a lower dose, or residual gain is negative. \\
Latent nonresponsive & Dosage response is nonresponsive, and the alpha path is flat, the stages stay low, or residual gain is low. \\
Decoder realization & Residual localization and decoder-score proxies look plausible, $S_1$ edit/center proxies are not low, but the exact decoded product is still wrong. \\
Residual template-like failure & Same-class/bucket residual control is both close to the true residual and relatively strong. \\
Latent high-risk untyped & Multiple high-risk proxies fire, but no specific latent rule has enough support; also used as the fallback when no rule fires. \\
\bottomrule
\end{tabular}
\end{table}

The endpoint primary label is chosen by evidence count with a fixed tie
priority, while all secondary fine labels are retained for auditing. The latent
primary priority is nonresponsive, program, residual/refinement, dosage,
decoder-realization, and then mixed/ambiguous; explicit best-alpha or
response-type evidence can promote dosage over residual refinement.

\subsection{Broader Impact and Asset Licenses}
\label{app:broader_impact}

\paragraph{Societal impact.}
This work aims to improve the inspectability of reaction models, which may support safer and more reliable use of machine-learning tools in chemical prediction and retrosynthesis. The main risk is indirect dual use: improved reaction-model analysis could be integrated into synthesis-planning workflows that might be used for unsafe or harmful chemical design. Our method is an audit-time diagnostic framework rather than an autonomous synthesis planner, and it should be used with expert chemical oversight.

\paragraph{Existing assets and licenses.}
We use the public USPTO-50K~\cite{schneider2016s} and USPTO-MIT~\cite{jin2017predicting} reaction benchmarks following prior work and cite the original dataset/source papers. Baseline numbers are taken from the cited papers under their reported USPTO-50K reaction-class-unknown settings. The transfer analyses in Appendix~\ref{app:rebuttal_analyses} use the public implementations of Graph2Edits~\cite{zhong2023graph2edits}, EditRetro~\cite{han2024iterative}, and Graph2SMILES~\cite{tu2022permutation} released by their authors, under their respective licenses. Molecular preprocessing and evaluation use RDKit, distributed under the BSD license. We do not use or release scraped proprietary datasets.
\subsection{Reference-Axis, Alignment, and Transfer Analyses}
\label{app:rebuttal_analyses}

This section gives the full results summarized in
Section~\ref{sec:axis_dependence}. Unless stated otherwise, results are
retrosynthesis top-1 exact-match accuracies (\%) on the USPTO-50K test split.

\textbf{Alignment-weight sweep.} Each $\lambda_{\mathrm{align}}$ setting in
Table~\ref{tab:lambda_sweep} is a separate checkpoint trained with the same
nominal training budget as the audited $\lambda_{\mathrm{align}}=0.1$
checkpoint. All other hyperparameters follow
Table~\ref{tab:urf_checkpoint_config_app}.

\textbf{Decoder fine-tuning.} Keeping the encoder and transformation frozen,
we fine-tune the decoder on paired inputs $[S_0\|g]\rightarrow
G_{\mathrm{target}}$. Because such pairs can train a new readout even when the
original decoder cannot execute $g$, and can also introduce an $S_0$ shortcut,
this intervention changes the representation-to-output interface rather than
the representation. Table~\ref{tab:decoder_finetune_app} scales the
displacement fed to the fine-tuned decoder.

\begin{table}[H]
\centering
\scriptsize
\setlength{\tabcolsep}{5pt}
\caption{Fine-tuned decoder, decoding $[S_0\|c\,g]$. Accuracy at $c=0$ exposes
the $S_0$ shortcut, so the $c=1$ value is a supervised-recoverability upper
bound. The original decoder gives $4.05\%$ at $c=0$ and $11.10\%$ at $c=1$.}
\label{tab:decoder_finetune_app}
\begin{tabular}{lrrrr}
\toprule
Displacement dosage $c$ & 0 & 0.25 & 0.5 & 1 \\
\midrule
Top-1 (\%) & 10.39 & 48.11 & 62.41 & 64.45 \\
\bottomrule
\end{tabular}
\end{table}

\textbf{Frozen-model linear calibration.} We fit a zero-intercept ridge map
$T(g)=gW$ on the training split so that $T(g)$ matches the realized
displacement $\Delta z$, using latent matching only (no output labels); the
model remains frozen. Mapped-axis decoding $S_0+T(g)$ rises from $11.10\%$ to
$21.25\%$. Re-decomposing $\Delta z$ around $T(g)$ gives $52.73\%$
residual-only versus $21.25\%$ parallel-only, a $31.48$-point gap, compared with
$53.07$ points around $g$.

\textbf{Matched-random null axes.} For each test reaction we draw eight random
unit axes $u$ with $\angle(u,\Delta z)=\angle(g,\Delta z)$. This matches the
parallel and residual component norms, and the distances of the
intervened states to $S_0$ and $S_1$; only the orientation changes. With the model frozen,
the target-defined axis yields a $53.07$-point Residual--Parallel separation
versus $37.13$ points for matched-random axes (difference $15.94$, 95\%
bootstrap CI $[14.83,17.10]$). Along the residual dosage path
$S_0+c\,\Delta z_\perp$, $c\in[0,1]$, the target-defined residual's accuracy
AUC is $6.61$ points higher than that of the matched-random residuals.

\textbf{Relative magnitude and target-axis sweep.} On the test split,
$\mathbb E[\|g\|/\|\Delta z\|]=0.9508\pm0.0016$ (s.e.; median $0.9553$), so
$S_1$ and $z_{\mathrm{target}}$ lie at comparable distances from $S_0$.
Table~\ref{tab:target_axis_sweep_app} decodes along the target axis directly.

\begin{table}[H]
\centering
\scriptsize
\setlength{\tabcolsep}{3pt}
\caption{Decoding $S_0+c\,g$ with the original frozen decoder. The response is
non-monotonic and peaks at $c=2.75$, far below $S_1$ ($64.41\%$).}
\label{tab:target_axis_sweep_app}
\begin{tabular}{lrrrrrrrrrrrrr}
\toprule
$c$ & 0 & 0.25 & 0.5 & 0.75 & 1 & 1.25 & 1.5 & 2 & 2.5 & 2.75 & 3 & 4 & 6 \\
\midrule
Top-1 & 4.05 & 4.59 & 6.55 & 8.59 & 11.10 & 13.40 & 15.20 & 17.74 & 18.99 & \textbf{19.53} & 19.29 & 17.52 & 11.72 \\
\bottomrule
\end{tabular}
\end{table}

\textbf{Graph2Edits transfer.} We train Graph2Edits~\citep{zhong2023graph2edits}
with its native edit loss plus an analogous endpoint-state alignment loss
($\lambda_{\mathrm{align}}=0.1$), adding no new parameters or decoder heads. The
resulting model reaches $55.44\%$/$83.74\%$ official end-to-end
retrosynthesis top-1/top-5. Because Graph2Edits acts through successive edits
rather than one global latent transition, we use an edit-level audit. For
4{,}999 test reactions we supply the ground-truth first edit, which prevents
earlier errors from contaminating the analysis. We take the recurrent states
immediately before ($S_0$) and after ($S_1$) that edit, define $z_{\mathrm{t}}$
by a fresh encoding of the target reactant graph, and decompose $S_1-S_0$
around $z_{\mathrm{t}}-S_0$. With all parameters frozen, the existing edit head
predicts the second edit from each intervened state
(Table~\ref{tab:graph2edits_app}).

\begin{table}[H]
\centering
\scriptsize
\setlength{\tabcolsep}{5pt}
\caption{Edit-level audit of endpoint-aligned Graph2Edits: second-edit top-1
(\%). Residual exceeds Parallel by $32.21$ points.}
\label{tab:graph2edits_app}
\begin{tabular}{lrrrrr}
\toprule
State & $S_0$ & Parallel & $z_{\mathrm{t}}$ & Residual & $S_1$ \\
\midrule
Top-1 (\%) & 0.68 & 16.58 & 22.74 & 48.79 & 75.80 \\
\bottomrule
\end{tabular}
\end{table}

\textbf{Post-hoc split-space alignment for other architectures.}
Table~\ref{tab:posthoc_alignment_app} repeats the split-aligned forward--retro
retrieval of Table~\ref{tab:audit_substrate} on the 5{,}007-reaction test set
for two external architectures. EditRetro recovers strong cross-direction
correspondence, whereas Graph2SMILES is much weaker, so post-hoc alignability
depends on the architecture and representation. This experiment tests
alignability only; it does not test whether $\Delta z_\perp$ is functionally
dominant in these models.

\begin{table}[H]
\centering
\scriptsize
\setlength{\tabcolsep}{5pt}
\caption{Post-hoc split-space forward--retro retrieval on $S_1$-type states.}
\label{tab:posthoc_alignment_app}
\begin{tabular}{lrr}
\toprule
Setting & AUC & R@1 \\
\midrule
UniRxnAxis split-aligned $S_1$ (Table~\ref{tab:audit_substrate}) & .9437 & .4068 \\
Graph2SMILES~\citep{tu2022permutation} & .7764 & .0184 \\
EditRetro~\citep{han2024iterative} & .9926 & .5866 \\
\bottomrule
\end{tabular}
\end{table}

\textbf{USPTO-MIT geometry.} We apply the audited checkpoint to USPTO-MIT
\citep{jin2017predicting} and recompute the per-sample geometry of
Table~\ref{tab:endpoint_geometry}. Absolute geometry shifts modestly, with
lower cosine alignment and larger orthogonal fractions, but the
forward--retro contrast is stable across datasets and splits
(Table~\ref{tab:uspto_mit_app}). This supports cross-dataset transfer of the
relative geometry, not of decoder executability.

\begin{table}[H]
\centering
\scriptsize
\setlength{\tabcolsep}{5pt}
\caption{Per-sample geometry on USPTO-50K and USPTO-MIT. F/R = forward/retro;
$\Delta$ = Retro $-$ Forward.}
\label{tab:uspto_mit_app}
\begin{tabular}{lrrrr}
\toprule
Dataset & Mean $\cos$ F/R & $\Delta\cos$ & Mean $r_\perp$ F/R & $\Delta r_\perp$ \\
\midrule
USPTO-50K test & .3282 / .6273 & +.2991 & .9322 / .7405 & $-$.1917 \\
USPTO-MIT train & .2561 / .5503 & +.2942 & .9529 / .7824 & $-$.1705 \\
USPTO-MIT val & .2560 / .5522 & +.2962 & .9531 / .7813 & $-$.1718 \\
USPTO-MIT test & .2566 / .5516 & +.2950 & .9528 / .7812 & $-$.1716 \\
\bottomrule
\end{tabular}
\end{table}

\newpage
\section*{NeurIPS Paper Checklist}

\begin{enumerate}

\item {\bf Claims}
    \item[] Question: Do the main claims made in the abstract and introduction accurately reflect the paper's contributions and scope?
    \item[] Answer: \answerYes{} 
    \item[] Justification: Yes. The abstract and introduction accurately reflect the paper’s main contribution: a directional latent audit framework for endpoint-supervised reaction models, with evidence that an endpoint-orthogonal residual is functionally important for retrosynthesis. We explicitly scope the claims as model-side audit rather than recovery of human chemical mechanisms or a deployable inference-time steering method, and state that the magnitude of the residual--parallel gap depends on the reference axis, the alignment weight, and the decoder interface (Section~\ref{sec:axis_dependence}). 

\item {\bf Limitations}
    \item[] Question: Does the paper discuss the limitations of the work performed by the authors?
    \item[] Answer: \answerYes{} 
    \item[] Justification: 
    Yes. Section~5 contains an explicit Limitations paragraph: the decomposition requires the ground-truth endpoint and is therefore an audit-time diagnostic; the gap magnitude depends on the reference axis, $\lambda_{\mathrm{align}}$, and the decoder interface; executable evidence covers retrosynthesis on USPTO-50K with two architectures (USPTO-MIT supports geometric transfer only); only deterministic top-1 is reported; and the reaction-center and failure labels are automatic proxies. Section~\ref{sec:axis_dependence} and Appendix~\ref{app:rebuttal_analyses} report the corresponding sensitivity analyses.

\item {\bf Theory assumptions and proofs}
    \item[] Question: For each theoretical result, does the paper provide the full set of assumptions and a complete (and correct) proof?
    \item[] Answer: \answerNA{} 
    \item[] Justification: This paper does not include theoretical result. 

    \item {\bf Experimental result reproducibility}
    \item[] Question: Does the paper fully disclose all the information needed to reproduce the main experimental results of the paper to the extent that it affects the main claims and/or conclusions of the paper (regardless of whether the code and data are provided or not)?
    \item[] Answer: \answerYes{} 
    \item[] Justification: The paper provides the information needed to reproduce the main experimental results that support the claims. The main text specifies the UniRxnAxis architecture, including the shared endpoint encoder, direction-conditioned transformation with task adapters, source-conditioned residual decoder, structured prediction loss, and latent endpoint-alignment loss. It also defines the directional decomposition used by the audit, including the source state $S_0$, transformed state $S_1$, encoded target endpoint $z_{\mathrm{target}}$, endpoint direction $g$, endpoint-parallel component $\Delta z_{\parallel}$, and endpoint-orthogonal residual $\Delta z_{\perp}$. The appendix provides the training and implementation  hyperparameters, audit-state definitions, aggregation domains, metric definitions, and intervention protocols needed to reproduce the staged decoding, residual replacement controls, decoder-boundary probes, dosage sweeps, atom-level diagnostics, and failure taxonomy. Appendix~\ref{app:rebuttal_analyses} specifies the reference-axis, alignment-weight, calibration, matched-random-axis, and cross-architecture/cross-dataset analyses. Code and processed data release details are addressed separately in the open-access checklist item.

\item {\bf Open access to data and code}
    \item[] Question: Does the paper provide open access to the data and code, with sufficient instructions to faithfully reproduce the main experimental results, as described in supplemental material?
    \item[] Answer: \answerYes{} 
    \item[] Justification: Code is publicly available at \url{https://github.com/yilish/UniRxnAxis}, and the main text gives this link in Section~4. The repository contains the UniRxnAxis implementation, the training and evaluation entry points, and the directional-audit scripts used for the staged decoding, residual replacement controls, decoder-boundary probes, dosage sweeps, atom-level diagnostics, and failure taxonomy, together with the environment specification and instructions for preparing the public USPTO-50K and USPTO-MIT benchmarks from their original sources. The paper also describes the data setting, splits, and evaluation protocol. 

\item {\bf Experimental setting/details}
    \item[] Question: Does the paper specify all the training and test details (e.g., data splits, hyperparameters, how they were chosen, type of optimizer) necessary to understand the results?
    \item[] Answer: \answerYes{}

\item[] Justification:
The paper specifies the experimental setting needed to understand the reported results. The main text describes the USPTO-50K evaluation setting, the reaction-class-unknown retrosynthesis benchmark, the model architecture, the structured endpoint prediction loss, the latent endpoint-alignment loss, and the directional decomposition used in the audit. The appendix provides the full experimental details, including data splits and preprocessing, model hyperparameters, optimizer and training configuration, checkpoint selection, audit-state definitions, metric definitions, aggregation domains, bootstrap protocol, and the intervention procedures used for staged decoding, residual replacement controls, decoder-boundary probes, dosage sweeps, atom-level diagnostics, and failure taxonomy.

\item {\bf Experiment statistical significance}
    \item[] Question: Does the paper report error bars suitably and correctly defined or other appropriate information about the statistical significance of the experiments?
    \item[] Answer: \answerYes{}

\item[] Justification:
The paper reports statistical uncertainty for the main experimental claims using bootstrap confidence intervals. The appendix explains that confidence intervals are computed by resampling test samples with replacement; edge-level quantities are first aggregated to sample-level means when sample identifiers are available, and aggregate-only quantities are explicitly marked rather than assigned invented intervals. Bootstrap intervals are reported for the staged decoding results, residual replacement controls, decoder-boundary and directional-derivative probes, dosage sweeps, atom-level diagnostics, failure-taxonomy statistics, the residual-dosage selector, and the matched-random-axis control in Appendix~\ref{app:rebuttal_analyses}. These intervals cover the experiments supporting the main claims about endpoint-parallel insufficiency, residual utility, residual robustness, decoder-level bond-edit effects, and failure diagnosis.

\item {\bf Experiments compute resources}
    \item[] Question: For each experiment, does the paper provide sufficient information on the computer resources (type of compute workers, memory, time of execution) needed to reproduce the experiments?
   \item[] Answer: \answerYes{}

\item[] Justification:
Table~\ref{tab:urf_checkpoint_config_app} reports the training hardware for the audited checkpoint: $4\times$ NVIDIA A10 GPUs for roughly 10 hours, about 40 GPU-hours. Each $\lambda_{\mathrm{align}}$ checkpoint in Table~\ref{tab:lambda_sweep} is trained under the same nominal budget, as stated in Appendix~\ref{app:rebuttal_analyses}. The remaining audit analyses are post-hoc evaluations of trained checkpoints and cost substantially less than training.
    
\item {\bf Code of ethics}
    \item[] Question: Does the research conducted in the paper conform, in every respect, with the NeurIPS Code of Ethics \url{https://neurips.cc/public/EthicsGuidelines}?
    \item[] Answer: \answerYes{} 
    \item[] Justification: We follow code of ethics of NeurIPS.


\item {\bf Broader impacts}
    \item[] Question: Does the paper discuss both potential positive societal impacts and negative societal impacts of the work performed?
   \item[] Answer: \answerYes{}

\item[] Justification:
The paper discusses societal impact in Appendix~\ref{app:broader_impact}. Positive impacts include improving the inspectability and reliability of machine-learning models for chemical reaction prediction and retrosynthesis, which may help chemists diagnose model failures and use predictive tools more cautiously. Potential negative impacts are indirect: better reaction-model auditing and prediction tools could contribute to computer-aided synthesis workflows that might be misused for harmful chemical design or unsafe synthesis planning if deployed without domain oversight. The work is primarily a model-side audit framework rather than a deployable synthesis-planning system, and we emphasize that its outputs should be treated as diagnostic evidence rather than autonomous chemical recommendations.
    
\item {\bf Safeguards}
    \item[] Question: Does the paper describe safeguards that have been put in place for responsible release of data or models that have a high risk for misuse (e.g., pre-trained language models, image generators, or scraped datasets)?
    \item[] Answer: \answerNA{} 
    \item[] Justification:The paper does not release a high-risk pretrained generative model, scraped dataset, image generator, language model, or autonomous synthesis-planning system. The experiments use the public USPTO-50K benchmark, and the proposed model is used as an audit-time diagnostic framework for analyzing reaction-model latent updates rather than as a deployed tool for autonomous chemical design. Therefore, the work does not require additional controlled-release safeguards of the type described in this checklist item.

\item {\bf Licenses for existing assets}
    \item[] Question: Are the creators or original owners of assets (e.g., code, data, models), used in the paper, properly credited and are the license and terms of use explicitly mentioned and properly respected?
    \item[] Answer: \answerYes{} 
    \item[] Justification:We use the public USPTO-50K reaction benchmark following prior retrosynthesis work and cite the original dataset/source papers. Baseline numbers are taken from the cited papers under their reported USPTO-50K reaction-class-unknown settings. We also use USPTO-MIT~\citep{jin2017predicting} and, for the transfer analyses in Appendix~\ref{app:rebuttal_analyses}, the public implementations of Graph2Edits~\citep{zhong2023graph2edits}, EditRetro~\citep{han2024iterative}, and Graph2SMILES~\citep{tu2022permutation} released by their authors, which we cite and use under their respective licenses. Molecular preprocessing and evaluation use RDKit, which is distributed under the BSD license. We do not use or release scraped proprietary datasets.

\item {\bf New assets}
    \item[] Question: Are new assets introduced in the paper well documented and is the documentation provided alongside the assets?
    \item[] Answer: \answerYes{} 
    \item[] Justification: The paper does not introduce a new dataset or benchmark; the experiments use existing public reaction benchmarks. The accompanying code release at \url{https://github.com/yilish/UniRxnAxis} is documented in the repository, which states its license and gives usage instructions for training, evaluation, and the directional audit.

\item {\bf Crowdsourcing and research with human subjects}
    \item[] Question: For crowdsourcing experiments and research with human subjects, does the paper include the full text of instructions given to participants and screenshots, if applicable, as well as details about compensation (if any)? 
    \item[] Answer: \answerNA{} 
    \item[] Justification: The paper does not involve crowdsourcing, user studies, surveys, behavioral experiments, or research with human subjects. All experiments are conducted on existing chemical reaction benchmark data and model-generated outputs.

\item {\bf Institutional review board (IRB) approvals or equivalent for research with human subjects}
    \item[] Question: Does the paper describe potential risks incurred by study participants, whether such risks were disclosed to the subjects, and whether Institutional Review Board (IRB) approvals (or an equivalent approval/review based on the requirements of your country or institution) were obtained?
    \item[] Answer: \answerNA{} 
    \item[] Justification: The paper does not involve crowdsourcing, user studies, surveys, behavioral experiments, or research with human subjects. All experiments are conducted on existing chemical reaction benchmark data and model-generated outputs, so IRB approval or equivalent human-subject review is not applicable.

\item {\bf Declaration of LLM usage}
    \item[] Question: Does the paper describe the usage of LLMs if it is an important, original, or non-standard component of the core methods in this research? Note that if the LLM is used only for writing, editing, or formatting purposes and does \emph{not} impact the core methodology, scientific rigor, or originality of the research, declaration is not required.
    \item[] Answer: \answerNA{} 
    \item[] Justification: The core method and experiments do not use LLMs as an important, original, or non-standard component. The paper proposes and evaluates a latent reaction-model audit framework based on UniRxnAxis, directional decomposition, decoder interventions, and benchmark evaluation. Any use of LLM tools, if any, was limited to writing, editing, or formatting assistance and did not affect the core methodology, scientific claims, experimental design, or results.

\end{enumerate}

\end{document}